\newcommand{\Task}[1]{\mathcal{Z}^{(#1)}}

 \newcommand{\Xt}[1]{\bm{X}^{(#1)}}
\newcommand{\Yt}[1]{\bm{Y}^{(#1)}}

\documentclass{article}
\usepackage{ijcai23}

\usepackage{times}
\usepackage{soul}
\usepackage{url}
\usepackage[hidelinks]{hyperref}
\usepackage[utf8]{inputenc}
\usepackage[small]{caption}
\usepackage{graphicx}
\usepackage{amsmath}
\usepackage{amsthm}
\usepackage{booktabs}
\usepackage{algorithm}
\usepackage{algorithmic}
\usepackage{bbm}
\usepackage{multirow}
\usepackage{amsfonts}
\usepackage{wrapfig}
\usepackage{hyperref}
\usepackage{url}
\usepackage{times}
\usepackage{helvet}
\usepackage{courier}
\usepackage{times}
\usepackage{epsfig}
\usepackage{graphicx}
\usepackage{amsmath}
\usepackage{amssymb}
\usepackage{url}
\usepackage{tabularx}
 \usepackage{verbatim}
\usepackage{bm}
\usepackage{subcaption}

 \usepackage{algorithm}

\usepackage{algorithmic}%

\title{Cognitively Inspired Learning of Incremental Drifting Concepts}

\author{Anonymous Submission}

\author{
Mohammad Rostami
\and
Aram Galstyan
\affiliations
University of Southern California
\emails
\{mrostami,galstyan\}@isi.edu
}

\begin{document}

\maketitle

\begin{abstract}

 Humans continually expand their learned knowledge to new domains and learn new concepts without any interference with past learned experiences. In contrast,   machine learning models perform poorly in a continual learning setting, where input data distribution changes over time. Inspired by the nervous system learning mechanisms, we develop a computational model that enables a deep neural network to learn new concepts and expand its learned knowledge to new  domains incrementally in a continual learning setting. We rely on the \textit{Parallel Distributed Processing} theory to encode abstract concepts in an embedding space in terms of a multimodal distribution. This embedding space is modeled by  internal data representations in a hidden network layer. We also leverage the \textit{Complementary Learning Systems theory} to equip the model with a memory mechanism to overcome catastrophic forgetting through implementing pseudo-rehearsal. Our   model   can generate pseudo-data points for experience replay  and  accumulate new experiences to past learned experiences without causing cross-task interference.

\end{abstract}

\section{Introduction}
\label{IntroductionUAI2021}
Humans continually abstract \textit{concept classes} from their input sensory  data to build semantic descriptions, and then update and expand these concepts as more experiences are accumulated~\cite{widmer1996learning}, and use them to express their ideas and communicate with each other~\cite{gennari1989models,lake2015human}. 
For example, ``cat'' and ``dog'' are   one of the first concept classes that many children learn to identify. Most humans     expand these concepts as \textit{concept drift} occurs, e.g., incorporating many atypical dog breeds into the ``dog'' concept, and also incrementally learn new concept classes, e.g. ``horse'' and ``sheep,'' as they acquire more experiences. Although this  concept   learning procedure occurs continually in humans, continual and incremental learning of concept classes remains a major challenge in artificial intelligence (AI). AI models are usually trained on a fixed number of classes and  the data distribution is   assumed to be stationary during model execution. Hence,   when an AI model is  trained or updated on  sequentially observed tasks with diverse   distributions or is trained on new classes, we generally need new annotated data points from the new classes ~\cite{rostami2018crowdsourcing} 
  and the model also would tend to forget what has been learned before due to cross-task interference, known as the phenomenon of \textit{catastrophic forgetting} ~\cite{french1991using}.

Inspired by the Parallel Distributed Processing (PDP) paradigm~\cite{mcclelland1986parallel,mcclelland2003parallel}, our goal is to enable a deep neural  network to learn \textit{drifting concept classes}~\cite{gama2014survey,rostami2023overcoming} incrementally and continually in a sequential learning setting.  PDP hypothesizes that abstract concepts are encoded in higher layers of the nervous system~\cite{mcclelland2003parallel,saxe2019mathematical}. Similarly, and based on behavioral similarities between artificial deep neural networks and the nervous system~\cite{morgenstern2014properties}
, we can assume that the data representations in  hidden layers of a deep network encode   semantic concepts with different levels of abstractions. We model these representations   as an embedding space in which semantic similarities between input data points are encoded in terms of geometric distances~\cite{jiang1997semantic}, i.e., data points that belong to the same concept class are mapped into separable clusters in the embedding space. When a new concept is abstracted, a new distinct cluster should be formed in the embedding space  to encode that new class. Incremental concepts learning is feasible by tracking and remembering the representation clusters that are formed in the embedding space and by considering their dynamics as more   experiences are accumulated in new   domains. 

We  benefit from the Complementary Learning Systems (CLS) theory~\cite{mcclelland1995there} to mitigate catastrophic forgetting.
CLS is based on empirical evidences that suggest  experience replay of recently observed patterns during sleeping and waking periods in the human brain helps to accumulate the new experiences to the past learned experiences without causing interference~\cite{mcclelland1995there,robins1995catastrophic}. According to this theory, hippocampus plays the role of a short-term memory buffer  that stores samples of recent experiences and  catastrophic forgetting is prevented by replaying samples from the hippocampal storage to implement pseudo-rehearsal in the neocortex during sleeping periods through enhancing past learned knowledge. Unlike AI memory buffers that store raw input data point, hippocampal storage  can only   store encoded  abstract representations.

 Inspired by the above two theories, we expand a base    neural   classifier with a decoder network, which is amended from a   hidden layer, to form an autoencoder with the hidden layer as its bottleneck.  The bottleneck is used to model the discriminative embedding space.  As a result of supervised learning, the embedding space becomes discriminative, i.e. a data cluster is formed for each concept class in the embedding space~\cite{mcclelland2003parallel,rostami2021transfer}.
  These clusters can be considered analogous to  neocortical representations in the brain, where the learned abstract concepts are encoded~\cite{mcclelland1986parallel}. We   use a multi-modal distribution to estimate this distribution~\cite{stan2021unsupervised,rostami2021lifelong}. We update this parametric distribution to accumulate new experiences to past learned experiences consistently.
Since our model is generative, we can  implement the offline memory replay process to prevent catastrophic forgetting~\cite{mcclelland1995there,rasch2013}.  When a new task arrives, we   draw random samples from the   multi-modal distribution and feed them into the decoder   to generate representative pseudo-data points. These pseudo-data points are then used to implement pseudo-rehearsal for experience replay~\cite{robins1995catastrophic}. 

\section{Related Work}
\label{RelatedWorkUAI2021}

\textbf{Continual learning:} the major challenge of continual learning  is  tackling catastrophic forgetting.  Previous works in the literature   mainly rely on  experience replay~\cite{li2018learning}. The core idea of experience replay is to implement pseudo-rehearsal by   replaying representative samples of past tasks along with the current task data to retain the learned distributions. Since storing these samples requires a memory buffer, the challenge is   selecting the representative samples to meet the buffer size limit.  For example, selecting uncommon samples that led to maximum effect in past experiences has been found to be effective~\cite{schaul2015prioritized}. However, as more tasks are learned,  selecting the effective samples becomes more complex.
The alternative approach is to use generative models that behave more similar to humans~\cite{french1999catastrophic}.  Shin et al.~(\cite{shin2017continual})   use  a generative adversarial structure  to mix the distributions of all tasks. It is also feasible to couple the distributions of all tasks in the bottleneck of  an  autoencoder~\cite{rostami2019Complementary,rostami2020generative}.  The shared distribution then can be used to generate  pseudo-samples~\cite{rannen2017encoder}.
 Weight consolidation using structural plasticity~\cite{lamprecht2004structural,zenke2017temporal,kirkpatrick2017overcoming} is   another approach to approximate experience replay.
The   idea is  to identify important weights that retain knowledge about a   task  and  then   consolidate them according to their relative importance for past tasks. Continual learning of sequential
tasks can be improved used high-level tasks descriptors to compensate for data scarcity~\cite{rostami2020using}.


\textbf{Incremental learning:} 
 forgetting in \textit{incremental learning}stems from updating the model when  new classes are incorporated, rather  concept drifts in a fixed number of learned classes. Hence, the goal is to learn new classes such that knowledge about the past learned classes is not overwritten.
A simple approach is to expand the base network as new classes are observed. Tree-CNN~\cite{roy2020tree} proposes a  hierarchical structure that grows like a tree when new classes are observed. The idea is to group new classes into feature-driven super-classes and find the exact label by limiting the search space. As the network grows, the new data can be used to train the expanded network. Sarwar et al.~\cite{sarwar2019incremental} add new convolutional filters in all layers to learn the new classes through new parameters.
The alternative approach  is to retain the knowledge about old classes in an embedding feature space.
Rebuffi et al.~\cite{rebuffi2017icaRL} proposed iCarl which maps images into a feature space that remains discriminative as more classes are learned incrementally. A fixed memory buffer is used to store exemplar images for each observed class. Each time a new class is observed, these images are used to learn a class-level   vector in the feature space such that the testing images can be classified using nearest neighbor with respect to these vectors. 

\textbf{Gaussian mixture model}:    are   useful for modeling   distributions that exhibit multiple modes or clusters. GMMs assume that the data is generated by a mixture of several Gaussian distributions, each representing a different cluster or mode in the data. The model is trained by estimating the parameters of the component Gaussians, including their means and variances, as well as the mixture weights that determine the relative contribution of each Gaussian to the overall distribution. GMMs are widely used in a variety of applications, including continual learning~\cite{rostami2019Complementary}.

\textbf{Contributions:} We develop a unified framework that addresses challenges of  both incremental learning and lifelong learning. Our idea is based on tracking and consolidating the multimodal distribution that is formed by the internal  data representations of sequential tasks in hidden layers of a neural network. We model this distribution as a  Gaussian mixture model (GMM) with time-dependent number of components. Concept drifts are learned by updating the corresponding GMM component for a particular class and new concepts are learned by adding new GMM components. We also make the model generative to implement experience replay. 

\section{Problem Statement}
\label{StatementUAI2021}

\begin{figure}[t!]
    \centering
    \includegraphics[width=\linewidth]{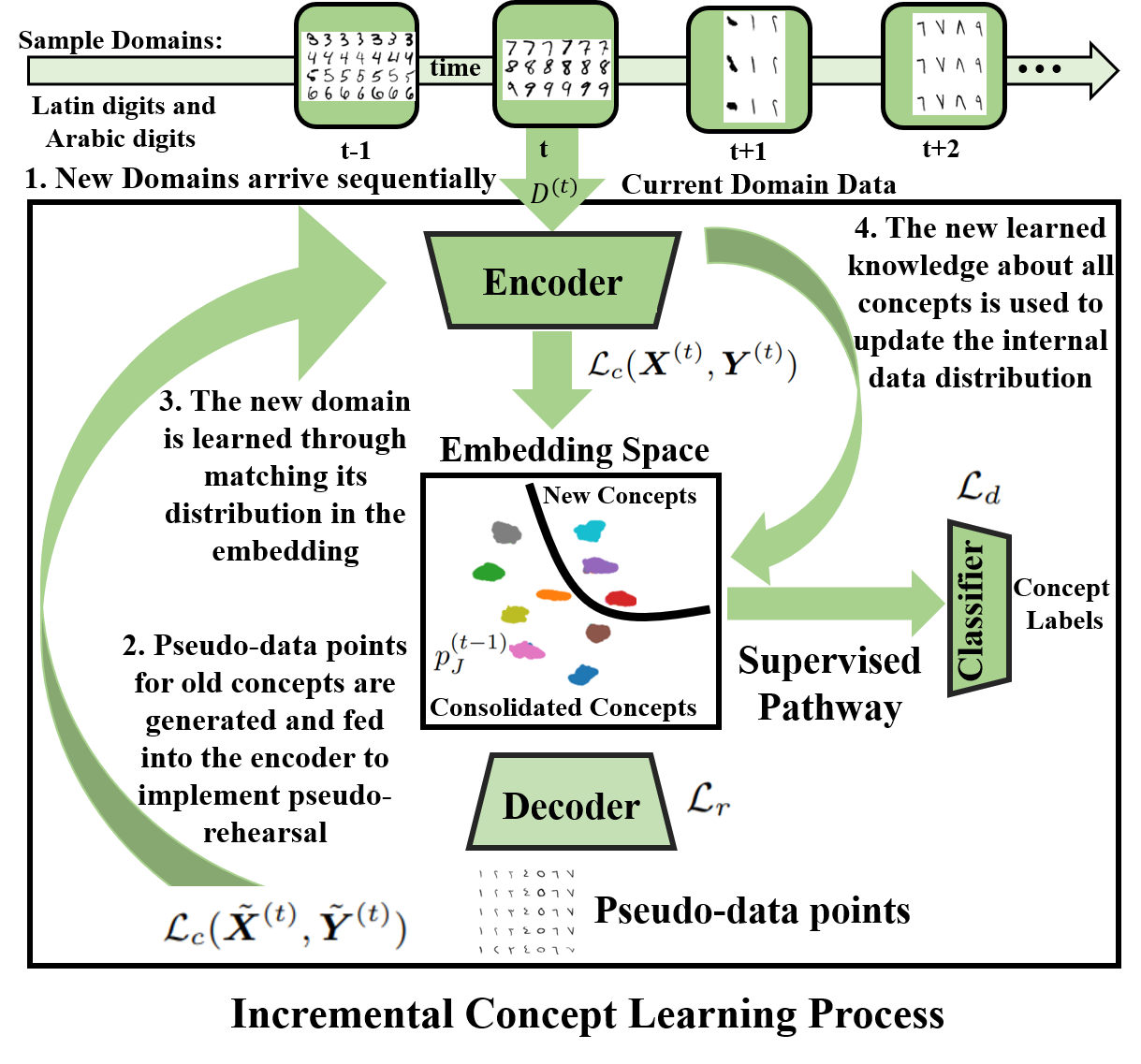}
         \caption{Block-diagram visualization of the proposed   Incremental Learning System. (Best viewed enlarged on screen and in color. Enlarged version is included in the Appendix) }
         \label{structureFigICML}
\end{figure}

Consider a learning agent which observes a sequence of observed tasks $\{\Task{t}\}_{t=1}^T$~\cite{chen2016lifelong} and after learning each task moves forward to learn the next task. Each task is a classification problem in a particular domain and each class represents a concept. The classes for each task can be new unobserved classes, i.e., necessitating incremental learning~\cite{rebuffi2017icaRL}, or drifted forms of the past learned classes, i.e., necessitating lifelong learning~\cite{chen2016lifelong},  or potentially a mixture of both cases. Formally, a task is characterized by a dataset $\mathcal{D}^{(t)} = \langle \Xt{t}, \Yt{t} \rangle$, where $  \Xt{t}=[\bm{x}_1^t,\ldots,\bm{x}_n^t]\in \mathbb{R}^{d \times n_t}$  and $ \Yt{t} \in \mathbb{R}^{k_t \times n_t}$ are the data points and one-hot labels, respectively. The goal is to train a time-dependent classifier function $f^{(t)}(\cdot):\mathbb{R}^d\rightarrow  \subset \mathbb{R}^{k_t}$- where $k_t$  is the number of classes for the $t$-th task and is fixed  for each task- such that the classifier continually generalizes on the tasks seen so far.  The data points  $\bm{x}_i^{(t)}\sim q^{(t)}(\bm{x})$ are assumed to be drawn i.i.d. from an unknown task distribution $q^{(t)}(\bm{x})$. Figure~\ref{structureFigICML} visualizes a   block-diagram of this continual and dynamic learning procedure. The agent needs to expand its knowledge about all the observed concepts such that it can  perform well on all the previous learned domains.

Learning each task in isolation is a standard supervised learning problem. After selecting a suitable parameterized family of functions $f_{\theta}^{(t)}:\mathbb{R}^{ d}\rightarrow \mathbb{R}^{k_t}$ with learnable  parameters $\theta$, e.g. a deep neural network with learnable  weight paramters $\theta$, we can solve for the optimal parameters using the empirical risk minimization (ERM): $\hat{ \theta}^{(t)}=\arg\min_{\theta}\hat{e}_{\theta}^{(t)}= \arg\min_{\theta}\sum_i \mathcal{L}_d(f_{\theta}^{(t)}(\bm{x}_i^{(t)}),\bm{y}_i^{(t)})$,   where $\mathcal{L}_d(\cdot)$  is a proper loss function. 
If $n_t$ is large enough, the  empirical risk expectation  would be a good approximation of the real expected risk function
 $e^{(t)}(\theta) = \mathbb{E}_{\bm{x}\sim q^{(t)}(\bm{x})}(\mathcal{L}_d(f_{\theta^{(t)}}(\bm{x}),f(\bm{x})))$. As a result, if the base parametric family is rich and complex enough for learning the task function, then the  ERM optimal model generalizes well on unseen test samples that are drawn from  $ q^{(t)}(\bm{x})$. 

For the rest of the paper, we consider the base model $f_{\theta^{(t)}}$ to be a deep neural network with an increasing output size  to encode incrementally observed classes. As stated, we rely on the  PDP paradigm. Hence, we    decompose the deep network  into an encoder sub-network $\phi_{\bm{v}}(\cdot):\mathbb{R}^{ d}\rightarrow \mathcal{Z}\subset \mathbb{R}^f$ with learnable parameter $\bm{v}$, e.g., convolutional layers of a CNN, and   a classifier sub-network $h_{\bm{w}}(\cdot)^{k_t}:\mathbb{R}^{ f}\rightarrow \mathbb{R}^{k_t}$ with learnable parameters $\bm{w}$, e.g., fully connected layers of a CNN, where $\mathcal{Z}$ denotes  the   embedding space in which the concepts will be be formed as separable clusters.

The concepts for each task are known a priori and hence
  new nodes are added to the   classifier sub-network output to incorporate the new classes at time $t$. We use a softmax layer as the last layer of the classifier subnetwork. Hence, we can consider the classifier to be a a maximum \textit{a posteriori} (MAP) estimator after training. This means that   the encoder network transforms the input data distribution into an internal multi-modal distribution with $k_t$ modes in the embedding space because  the embedding space $\mathcal{Z}$ should be concept-discriminative for good generalization. Each concept class is represented by a single mode of this distribution. We    use a Gaussian mixture model (GMM) to model and approximate this distribution  (see Figure~\ref{structureFigICML}, middle panel).  Catastrophic forgetting is the result of changes in this internal distribution when changes in the input distribution leads to updating the internal distribution heuristically. Our idea is to track changes in the data distribution and update and consolidate the internal distribution such that the acquired knowledge from past experiences is not overwritten when learning new tasks. 
 
 The main challenge is   to adapt the   network $f_{\theta}^{(t)}(\cdot)$ and the standard ERM training loss  such that we can track the internal distribution continually and accumulate the new acquired knowledge consistently to the past learned knowledge with minimum interference. For this purpose, we form a generative model by  amending the base model with a  decoder $\psi_{\bm{u}}: \mathcal{Z}\rightarrow \mathbb{R}^d$, with learnable parameters $\bm{u}$. This decoder maps back the internal representations to reconstruct the input data point in the input space such that the pair $(\phi_{\bm{u}},\psi_{\bm{u}})$ forms an autoencoder. 
 According to our previous discussion,  a multi-modal distribution would be formed in the bottleneck of the autoencoder upon learning each task.  This distribution encodes the learned knowledge about the concepts that have been learned from past experiences so far.  
 If we   approximate this distribution with a GMM, we can generate pseudo-data points that represent the previously learned concepts   and use them for pseudo-rehearsal. For this purpose, we can simply draw   samples from all modes of the GMM and feed these samples into the decoder subnetwork  to generate a pseudo-dataset (see Figure~\ref{structureFigICML}).
After learning each task, we can update the GMM  estimate  such that the new knowledge acquired  is accumulated to the past gained knowledge consistenly to avoid interference. By doing this procedure continually, our model is able to learn drifting concepts incrementally.  Figure~\ref{structureFigICML} visualizes  this     repetitive procedure in this lifelong learning setting.

\section{Proposed Algorithm}
\label{SolutionUAI2021}

 When the first task  is learned, there is no prior experience and hence  learning    reduces   the following:
 \begin{equation}
{\small
\begin{split}
&\min_{\bm{v},\bm{w},\bm{u}} \mathcal{L}_{c}(\bm{X}^{(1)},\bm{Y}^{(1)}) =\min_{\bm{v},\bm{w},\bm{u}} \frac{1}{n_1}\sum_{i=1}^{n_1} \Bigg(  \mathcal{L}_d\Big(h_{\bm{w}}(\phi_{\bm{v}}\big(\bm{x}_i^{(1)})\big),\bm{y}_i^{(1)}\Big)\\&+\gamma \mathcal{L}_r\Big(\psi_{\bm{u}}\big(\phi_{\bm{v}}(\bm{x}_i^{(1)})\big),\bm{x}_i^{(1)}\Big)\Bigg),
\end{split}
}
\label{eq:mainINCconceptIncremental}
\end{equation}  
\normalsize
where  $\mathcal{L}_d$ is the   discrimination loss, e.g., cross-entropy loss,  $\mathcal{L}_r$ is the reconstruction loss   for the autoencoder, e.g., $\ell_2$-norm,  $\mathcal{L}_c$ is the combined loss, and $\gamma$ is a trade-off parameter  between the   terms. 
When the first task is learned, also any future task, according to the PDP hypothesis, a multi-modal distribution $p^{(1)}(\bm{z})=\sum_{j=1}^{k_1}\alpha_j\mathcal{N}(\bm{Z}|\mu_j,\Sigma_j)$  with $k_1$ components is formed in the embedding space.  We   assume that this distribution can be modeled with a GMM. Since the labels for the input task data samples are known, we  use MAP estimation to  recover the GMM parameters (see Appendix for details). 
 Let $\hat{p}^{(1)}(\bm{z})$ denotes the estimated   distribution.

 \begin{algorithm}
\caption{$\mathrm{ICLA}\left (\lambda, \gamma, \tau \right)$\label{ICMLalgorithm}} 
 {\small
\begin{algorithmic}[1]
\STATE \textbf{Input:} labeled training datasets in a sequence
\STATE\hspace{8mm} $\mathcal{D}^{(t)}=(\{\bm{X}^{(t)},\bm{X}^{(t)})\}$ for $t=\ge 1$
\STATE \textbf{Initial   Learning}: learn the first task   via Eq.~\eqref{eq:mainINCconceptIncremental}
\STATE \textbf{Fitting GMM:}
\STATE \hspace{8mm}estimate $\hat{p}_{J}^{(1)}(\cdot)$ using $ \{\phi_{\bm{v}}(\bm{x}_i^{(1)}),\bm{y}_i^{(1)}\}_{i=1}^{n_t}$
\STATE{\textbf{For} $t \ge 2$ }
\STATE \hspace{3mm} \textbf{Generate the pseudo dataset:}
\STATE \hspace{8mm}   $\tilde{\mathcal{D}}^{(t)}  = 
\{( \tilde{\bm{x}}_{i}^{(t)}=\psi(\tilde{\bm{z}}_{i}^{(t)}), \tilde{\bm{y}}_{i}^{(t)})\}$
\STATE \hspace{8mm}$(\tilde{\bm{z}}_{i}^{(t)},\tilde{\bm{y}}_{i}^{(t)})\sim \hat{p}^{(t-1)}(\cdot) $
\STATE \hspace{3mm}\textbf{Task learning:} 
\STATE \hspace{8mm}learnable parameters are updated  via Eq.~\eqref{eq:mainINCconceptnextIncremental}
\STATE \hspace{3mm}\textbf{Estimating the internal distribution:}
\STATE \hspace{8mm}update $\hat{p}^{(t)}(\cdot)$  with $k_{Tot}^{(t)}$ components via the 
\STATE \hspace{8mm}combined samples  $ \{\phi_{\bm{v}}(\bm{x}_i^{(t)}),\phi_{\bm{v}}(\tilde{\bm{x}}_{i}^{(t)})\}_{i=1}^{n_t}$
\STATE \textbf{EndFor} 
\end{algorithmic}}
\end{algorithm} 
 
As subsequent tasks are learned,  the internal distribution should be updated continually to accumulate the new acquired knowledge.
Let $k_t=k_{old}^t+k_{new}^t$, where $k_{old}^t$ denotes the number of   the previously learned concepts that exist in the current task and $k_{new}^t$ denotes the number of the new observed classes. 
Hence, the total number of learned concepts   until $t=T$ is $k_{Tot}^{T}=\sum_{t=1}^{T}k_{new}^t$. Also, let the index set $\mathbb{N}^T_{Tot}= \{1,\ldots, k_{Tot}^{T}\}$ denotes an order on the  classes $C_j$, with $j \in \mathbb{N}^T_{Tot}$, that are observed until $t=T$.  Let $\mathbb{N}_T=\mathbb{N}_{old}^T\cup \mathbb{N}_{new}^T=\{i_1,\ldots, i_{k_T}\}\subset \mathbb{N}^T_{Tot}$ contains the $k_T$  indices of the existing concepts   in 
  $\Task{T}$. To update the internal distribution after learning $\Task{T}$, the number of distribution modes should be updated to $k_{Tot}^{T}$. Additionally, catastrophic forgetting must be mitigated using experience replay. We can draw random samples from the GMM distribution $\bm{z}_i\sim\hat{p}^{(T-1)}(\bm{z})$  and then pass each  sample through the decoder $\psi(\bm{z}_i)$  to generate pseudo-data points for pseudo-rehearsal.
 Since each particular concept is represented by exactly one mode of the internal GMM distribution, the corresponding pseudo-labels for the generated pseudo-data points are known. Moreover, the confidence levels for these labels are also known from the classifier softmax layer. 
 To generate a clean pseudo-dataset, we can set a threshold $\tau$ and only pick the pseudo-data points   for which the model confidence level   is more than $\tau$.
We also  generate a balanced pseudo-dataset with respect to the learned classes. Doing so, we ensure suitability of a GMM with $k_{Tot}^{T}$ components   to estimate the empirical distribution accurately after learning the next tasks. 
 
Let $\mathcal{\tilde{D}}^{(t)} = \langle \psi(\tilde{\bm{Z}}^{(t)}), \tilde{\bm{Y}}^{(t)} \rangle$ 
 denotes the pseudo-dataset, generated at time $t$ after learning the tasks $\{\Task{s}\}_{s=1}^{t-1}$. We form the following   objective    to  learn the task $\Task{t}$, $\forall t\ge 2$:
\begin{equation}
{\small
\begin{split}
&\min_{\bm{v},\bm{w},\bm{u}} \mathcal{L}_{c}(\bm{X}^{(t)},\bm{Y}^{(t)})+\mathcal{L}_{c}(\tilde{\bm{X}}^{(t)},\tilde{\bm{Y}}^{(t)})+\\&\lambda \sum_{j\in \mathbb{N}_{old}^t } D\Big(\phi_{\bm{v}}(q^{(t)}(\bm{X}^{(t)})|C_j),\hat{p}^{(t-1)}(\tilde{\bm{Z}}^{(t)})|C_j)\Big), 
\end{split}
}
\label{eq:mainINCconceptnextIncremental}
\end{equation}    
where $D(\cdot,\cdot)$ is a  probability    metric and $\lambda$ is a  parameter.

The first and the second terms in Eq.~\eqref{eq:mainINCconceptnextIncremental} are combined loss terms for  the current task training dataset and the generated pseudo-dataset that represent the past tasks, defined similar to Eq.~\eqref{eq:mainINCconceptIncremental}. 
 The second term in Eq.~\eqref{eq:mainINCconceptnextIncremental} mitigates catastrophic forgetting through pseudo-rehearsal process. The third  term is a crucial term to guarantee that our method will  work in a lifelong learning setting. This term enforces that each concept is encoded in one mode of the internal distribution across all tasks. This term is computed on the subset of the concept classes that are shared between the current task   and the pseudo-dataset, i.e, $\mathbb{N}^t_{old}$, to enforce   consistent knowledge accumulation. Minimizing the probability metric $D(\cdot,\cdot)$   enforces that the internal conditional distribution for the current task $\phi_{\bm{v}}(q^{(t)}(\cdot|C_j))$, conditioned on a particular shared concept $C_j$, to be close to the   conditional shared distribution $p^{(t-1)}(\cdot|C_j)$. Hence, both form a single mode of the internal distribution and concept drifting is mitigated. Conditional matching of the two distributions is feasible as we have access to pseudo-labels.  Adding this term guarantees that we can continually use a GMM with exactly $k_{Tot}^{(t)}$ components  to capture the   internal distribution in this lifelong learning setting. 
The remaining task is to select a suitable probability   metric $D(\cdot,\cdot)$   for solving  Eq.~\eqref{eq:mainINCconceptnextIncremental}.  Wasserstein Distance (WD) metric has been found to be an effective choice for deep learning due to its applicability for gradient-based optimization~\cite{courty2017optimal}. To reduce the computational burden of computing WD,   we use the Sliced Wasserstein Distance (SWD)~\cite{bonneel2015sliced}. 
(for   details on the SWD,   refer to the Appendix). Our  Incremental  Concept Learning Algorithm (ICLA) method  is summarized in Algorithm~\ref{ICMLalgorithm}.

\section{Theoretical Analysis}
\label{TheoreticalUAI2021}

We demonstrate that ICLA   minimizes an upperbound for the expected risk of the learned concept classes across all the previous tasks  for all $t$.  
We perform our analysis in the embedding space as an input space and consider   the   hypothesis class     $\mathcal{H}=\{h_{\bm{w}}(\cdot)|h_{\bm{w}}(\cdot):\mathcal{Z}\rightarrow\mathbb{R}^k_t, \bm{w} \in \mathbb{R}^H\}$.
Let $e_{t}(\bm{w})$  denote the real risk for 
 a given function $h_{\bm{w}^{(t)}}(\cdot)\in\mathcal{H}$  when  used on task $\Task{t}$  data representations in the embedding space. Similarly,  $\tilde{e}_{t}(\bm{w})$ denotes the observed risk for the  function $h_{\bm{w}^{(t)}}(\cdot)$ when   used on the pseudo-task, generated by sampling the learned GMM distribution   $\hat{p}^{(t-1)}$. 
 Finally, let $e_{t,s}(\bm{w})$ denote the risk of the model $h_{\\bm{w}^{(t)}}(\cdot)$ when used only on the concept classes in the set $\mathbb{N}_{s}\subset \mathbb{N}^t_{Tot}$, for $ \forall s\le t$, i.e., task specific classes, after learning the task   $\Task{t}$. 

\textbf{Theorem~1 }: Consider two tasks $\Task{t}$ and $\Task{s}$, where $s\le t$.  Let $h_{\bm{w}^{(t)}}$   be an optimal classifier trained for the $\Task{t}$ using the ICLA algorithm. Then for any $d'>d$ and $\zeta<\sqrt{2}$, there exists a constant number $N_0$ depending on $d'$ such that for any  $\xi>0$ and $\min(\tilde{n}_{t|\mathbb{N}_{s}},n_{s})\ge \max (\xi^{-(d'+2),1})$ with probability at least $1-\xi$ for  $h_{\bm{w}^{(t)}}\in \mathcal{H}$, then:
\begin{equation}
\small
\begin{split}
e_{s}(\bm{w})\le & e_{t-1,s}(\bm{w}) +W(\hat{p}^{(t-1)}_{s}, \phi(\hat{q}^{(s)}))+e_{\mathcal{C}}(\bm{w}^*)+ \\&\sqrt{\big(2\log(\frac{1}{\xi})/\zeta\big)}\big(\sqrt{\frac{1}{\tilde{n}_{t|\mathbb{N}_{s}}}}+\sqrt{\frac{1}{n_{s}}}\big),
\end{split}
\label{eq:theroemfromcourtyCatForIncremental}
\end{equation}    
where $W(\cdot,\cdot)$ denotes the WD metric, $\tilde{n}_{t|\mathbb{N}_{s}}$ denotes  the pseudo-task samples that belong to the classes in $\mathbb{N}_{s}$, $\phi(\hat{q}^{(s)}(\cdot))$ denotes the empirical marginal distribution for  $\Task{s}$ in the embedding,   $\hat{p}^{(t-1)}_{s}$
is the conditional empirical shared distribution when the distribution $\hat{p}^{(t-1)}(\cdot)$ is conditioned to the classes in $\mathbb{N}_s$, and $e_{\mathcal{C}}(\bm{w}^*)$ denotes the optimal model learned for the combined risk of the   tasks on the shared classes in $\mathbb{N}_s$, i.e., $\bm{w}^*= \arg\min_{\bm{w}} e_{\mathcal{C}}(\theta)=\arg\min_{\bm{w}}\{ e_{t,s}(\bm{w})+  e_{s}(\bm{w})\}$. This is a model with the best performance if the tasks could be learned simultaneously.

\textit{\textbf{Proof}}:  included in  the Appendix due to page limit.

We then use Theorem~1  to conclude the following lemma:

\textbf{Lemma~1 }: Consider the ICLA algorithm     after  learning $\Task{T}$. Then all tasks $t<T$ and under the conditions of Theorem~1, we can conclude the  following inequality:
\begin{equation}
\small
\begin{split}
&e_{t}(\bm{w})\le  e_{T-1,t}(\bm{w})+ W(\phi(\hat{q}^{(t)}), \hat{p}_{t}^{(t)})+e_{\mathcal{C}}(\bm{w}^*)+\\&\sum_{s=t}^{T-2} W(\hat{p}_{t}^{(s)}, \hat{p}_{t}^{(s+1)}) +\sqrt{\big(2\log(\frac{1}{\xi})/\zeta\big)}\big(\sqrt{\frac{1}{n_t}}+\sqrt{\frac{1}{\tilde{n}_{t|\mathbb{N}_t}}}\big),
\end{split}
\label{eq:theroemfromcourtyCatForoursIncremental}
\end{equation}    
\textit{Proof}:  included in  the Appendix due to page limit.


Lemma~1 concludes that when  a new task is learned at time $t=T$, ICLA updates the model parameters  conditioned on minimizing  the upper bound of $e_{t}$ for all $t < T$ in Eq.~\ref{eq:theroemfromcourtyCatForoursIncremental}. The last term in Eq.~\ref{eq:theroemfromcourtyCatForoursIncremental} is a small constant term when the number of training data points is large. If the network is complex enough so that the PDP hypothesis holds, then the classes would be separable in the embedding space and in the presence of enough labeled samples, the terms  $e_{T-1,t}(\bm{w})$ would be small because  $e_{T-1}(\bm{w})$ is minimized using ERM. 
 The term $W(\phi(\hat{q}^{(t)}), \hat{p}_{t}^{(t)})$ would be small because we deliberately fit  
the GMM distribution $\hat{p}^{(t)}$ to the distribution  $\phi(\hat{q}^{(t)})$ in the embedding space when learning the task $\Task{t}$.  Existence of this term indicates that our algorithm requires   that internal distribution can be fit with a GMM distribution with high accuracy and this limits applicability of our algorithm. Note however, all parametric learning algorithms face this limitation. 
The term  $e_{\mathcal{C}}(\bm{w}^*)$ is small because we continually match the distributions in the embedding space class-conditionally. Hence, if the model is trained on task $\Task{t}$ and the pseudo-task at $t-T$, it will perform well on both tasks. Note that this is not  trivial because if the wrong classes are matched across the domains in the embedding space, the term $e_{\mathcal{C}}(\bm{w}^*)$ will not be minimal. Finally, the sum term in
Eq.~\ref{eq:theroemfromcourtyCatForoursIncremental} indicates the effect of experience replay. Each term in this sum is minimized at $s=t+1$  because we draw random samples from $\hat{p}_{t}^{(t)}$ and then train the autoencoder to enforce    $\hat{p}_{t}^{(t)} \approx \psi(\phi(\hat{p}_{t}^{(t)}))$. Since all the terms in the upperbound of $e_{t}(\bm{w})$  in Eq.~\ref{eq:theroemfromcourtyCatForoursIncremental} are minimized when a new task is learned, catastrophic forgetting of the previous tasks  will be mitigated. Another important intuition from Eq.~\ref{eq:theroemfromcourtyCatForoursIncremental} is that as more 
tasks are learned after learning a task, the upperbound becomes looser as more terms are accumulated in the sum which enhances forgetting. This observation accords with our intuition about forgetting as more time passes after initial learning time of a task or concept.

\section{Experimental Validation}
\label{ExperimentalUAI2021}
 To the best of knowledge, no prior method has been developed to address challenges of both continual and incremental learning setting at the same time. For this reason, we validate  our method on two   sequential task learning settings:   incremental learning and  continual incremental learning. Incremental learning is a special case of our learning setting  when each concept class is observed only in one task and concept drift does not exist. We use this special case to compare our method against existing incremental learning approaches in the literature to demonstrate that our method is comparably effective.
 Our implementation is available as a supplementary.
 
\textbf{Evaluation Methodology}: We use the same  network structure for all the methods for fair comparison. 
 To visualize the results, we generate learning curves by plotting the model performance on the testing split of datasets versus the training epochs, i.e, to model  time.  We report  the average performance  of five runs. Visualizing learning curves allows studying temporal aspects of learning.  For comparison, we provide learning curves for: (a)  full experience replay (FR) which  stores the whole training data for all the previous tasks and (b) experience replay using a memory buffer (MB) with a fixed size, similar to Li et. al~(\cite{li2018learning}). 
 At each time-step, the buffer  stores an equal number of samples per concept from the previous tasks. When a new task is learned, a portion of old stored samples are discarded and replaced with samples from the new task to keep the buffer size fixed. FR serves as a best achievable upperbound to measure the  effectiveness of our method against the upperbound.  For  more    details about the experimental setup and all parameteric values, please refer to the Appendix and the provided code.

\subsection{Incremental Learning}
The   classes   are encountered only at one task in incremental learning.
We design two incremental learning experiments using the MNIST and the Fashion-MNIST datasets. Both datasets are classification datasets with ten classes. MNIST dataset consists of gray scale images of handwritten digits and Fashion-MNIST  consists of images of common fashion products. We consider an incremental learning setting with nine tasks for the MNIST dataset. The first task is a binary classification of digits $0$ and $1$ and each subsequent task involves learning a new digit. 
 The   setup for Fashion-MNIST dataset is similar, but  we  considered four tasks and  each   task involves learning two  fashion classes. We use a memory buffer with the fixed size of 100 for MB. We build an autoencoder by expanding  a VGG-based  classifier by mirroring the layers. 

\begin{figure}[tb!]
    \centering
           \begin{subfigure}[b]{0.23\textwidth}\includegraphics[width=\textwidth]{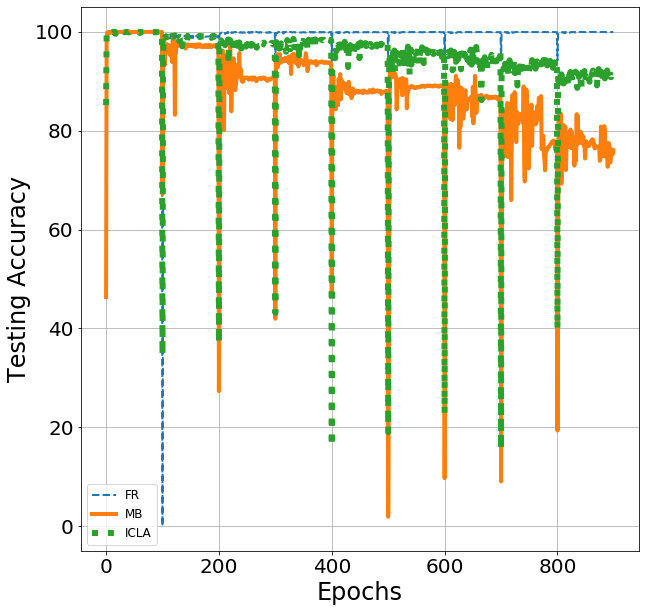}
           \centering
        \caption{MNIST}
        \label{ICMLDALfig:MNISTUSPS}
    \end{subfigure}
    \begin{subfigure}[b]{0.23\textwidth}\includegraphics[width=\textwidth]{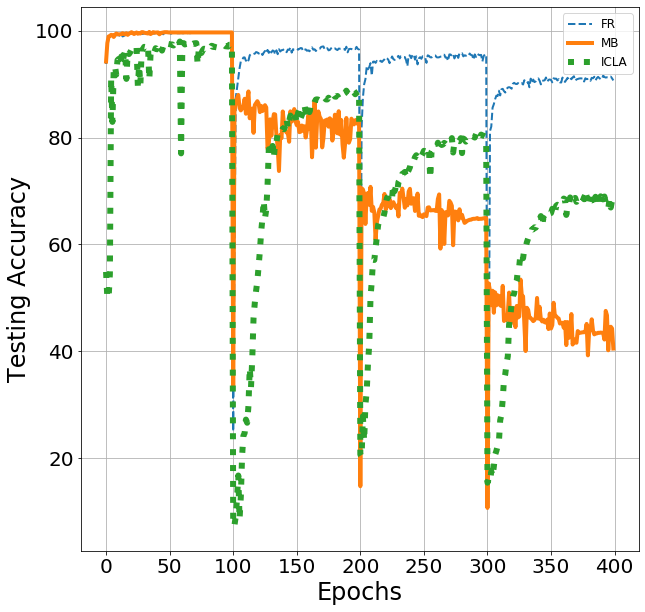}
           \centering
        \caption{FMNIST }
        \label{ICMLDALfig:USPSMNIST}
    \end{subfigure}
     \caption{Learning curves  for the   incremental learning experiments  (a)   MNIST     and (b) Fashion-MNIST (FMNIST) datasets; (c) MNIST performance comparison (Best viewed in color on screen. See Appendix for   enlarged versions.)  }\label{ICMLDALfig:resultsCatforgetRelated}
\end{figure}

Figure~\ref{ICMLDALfig:resultsCatforgetRelated} presents  results for the   designed experiments. For simplicity, we have provided condensed results for all tasks in a single curve. Each task is learned in 100 epochs and
at each epoch, the model performance is computed  as the average classification rate over all the classes,   observed before. We report performance on the standard testing split of  each dataset for the observed classes.
Figure~\ref{ICMLDALfig:MNISTUSPS} and present the learning curves for the MNIST experiments. Similarly, Figure~\ref{ICMLDALfig:USPSMNIST}    present  learning curves for the Fashion-MNIST experiments.
We can see in both figures that  FR (dashed blue curves) leads to  superior performance. This is according to expectation but as we discussed, the challenge is the requirement for a memory buffer with an unlimited size. The buffer cannot have a fixed size as the number of data points grows  when more tasks are learned. MB (solid yellow  curves)
is   initially somewhat effective and comparable with ICLA, but as more tasks are learned, forgetting effect becomes more severe. 
This is because fewer data points per task can be stored in the buffer with fixed size as more tasks are learned. As a result, the stored samples would not be sufficiently representative of the past learned tasks. In comparison, we can generate as  many pseudo-data points as desired. 

We can also see in Figure~\ref{ICMLDALfig:MNISTUSPS}  and Figure~\ref{ICMLDALfig:USPSMNIST} that  ICLA (dotted green curves)  is able to mitigate   catastrophic forgetting considerably better than MB and the performance difference between ICLA and MB increases as more tasks are learned. We also observe that ICLA is more effective for MNIST dataset. This is  because FMNIST data points are more diverse. As a result, generating pseudo-data points that look more similar to the original data points is easier for the MNIST dataset given that we are using the same network structure for both tasks.
Another   observation is that the major performance degradation for ICLA occurs  each time the network starts to learn a new concept class as initial sudden drops. This degradation occurs due to the existing distance between the distributions $\hat{p}_{J,k}^{(T-1)}$ and $\phi(q^{(s)})$ at $t=T$ for $s < T$. Although ICLA minimizes this distance, the autoencoder is not   ideal and this distance is non-zero in practice.
 
 For comparison against existing works, we   have listed our performance and a number of  methods for incremental learning on MNIST    in Table~\ref{table:continualtabDA2}. Two sets of tasks for incremental learning setting have been designed using MNIST in the literature: 5 tasks (5T)  setting and 2 tasks (2T) setting. In the 2T setting, two tasks are define involving digits $(0-4)$ and $(5-9)$. In the 5T setting, five binary classification tasks are defined involving digits $(0,1)$ to $(8,9)$.  We have compared our performance against several   methods, representative of prior works:  CAB~\cite{he2018overcoming}, IMM~\cite{lee2017overcoming},   OWM~\cite{zeng2019continual},   GEM~\cite{lopez2017gradient}, iCarl~\cite{rebuffi2017icaRL},  GSS~\cite{aljundi2019gradient}, DGR~\cite{shin2017continual}, and MeRGAN~\cite{wu2018memory}.  The CAB, IMM, and OWM methods are based on regularizing the network weights. The GEM, iCarl, and GSS methods use a memory buffer to store selected samples. Finally,   DGR and MeRGAN methods are  based on generative replay similar to ICLA but use adversarial learning.  We have reported the classification accuracy on the ten digit classes after learning the last task in Table~\ref{table:continualtabDA2}. A memory buffer with a fixed size of 100 is used for GEM, iCarl, and GSS. Following these works, an MLP  with two layers  is used as the base model for fair comparison.

We observe in Table~\ref{table:continualtabDA2} that when the buffer size is small, buffer-based methods  perform poorly. Methods based   on weight regularization   perform quite well but note that these methods limit the network learning capacity. As a result, when the number of tasks grow, the network cannot be used to learn new tasks. 
Generative methods, including ICLA, perform better compared to buffer-based methods and at the same time do not limit the network learning capacity  because the network weights can   change after generating the pseudo-dataset. Although  ICLA has the state-of-the-art performance for these tasks,  there is no superior method for all conditions, because by changing the experimental setup, e.g., network structure, dataset, hyper-parameters such as memory buffer, etc, a different method may have the best performance result.  However, we can conclude that ICLA has a superior performance when the network size is small and using a memory buffer is not possible, i.e., we have limited learning resources.

  \begin{table}[tb!]
 \centering 
{  \scriptsize
\begin{tabular}{lc|cc}   
\multicolumn{2}{c}{Method}    &  2T & 5T \\
\hline
\multicolumn{2}{c|}{CAB~\cite{he2018overcoming}}&  94.9$\pm$0.3 & - \\ 
\multicolumn{2}{c|}{IMM~\cite{lee2017overcoming}}  &   94.1$\pm$0.3& -  \\
\multicolumn{2}{c|}{OWM~\cite{zeng2019continual}}  &   96.3$\pm$0.1& -  \\
\hline
\multicolumn{2}{c|}{GEM~\cite{lopez2017gradient}}   &  - & 78.0  \\ 
\multicolumn{2}{c|}{iCarl~\cite{rebuffi2017icaRL}}  & - & 81.0 \\
\multicolumn{2}{c|}{GSS~\cite{aljundi2019gradient}}   & -  & 61.0\\
\hline
\multicolumn{2}{c|}{DGR~\cite{shin2017continual}}&    88.7$\pm$2.6& -  \\
\multicolumn{2}{c|}{MeRGAN~\cite{wu2018memory}}&    97.0& -  \\
\hline
\hline
\multicolumn{2}{c|}{ICLA}&    97.2$\pm$0.2 & 91.6$\pm$0.4   \\ 
\hline
\end{tabular}}
\caption{ Classification accuracy for MNIST.}

\label{table:continualtabDA2}
\end{table}

\begin{figure*}[tb!]
    \centering
           \begin{subfigure}[b]{.245\textwidth}\includegraphics[width =\textwidth,]{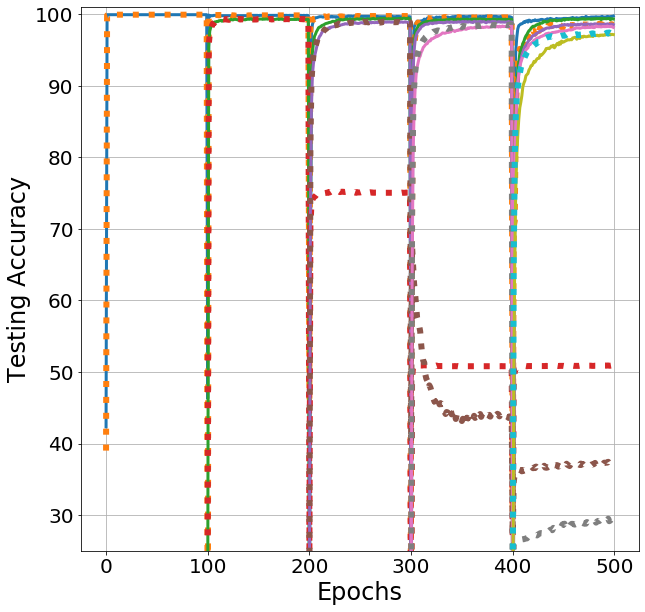}
           \centering
        \caption{FR vs. MB}
        \label{ICMLDALfig:BPvsCLEER}
    \end{subfigure}
    \begin{subfigure}[b]{.245\textwidth}\includegraphics[width=\textwidth]{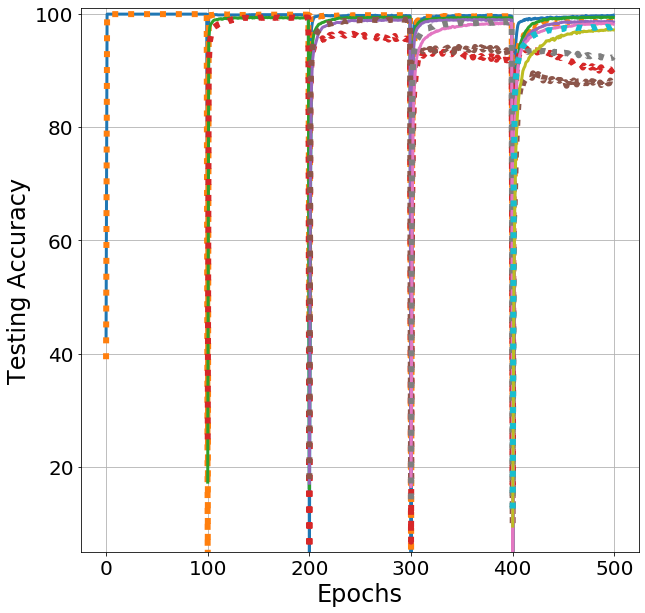}
           \centering
        \caption{FR vs. ICLA}
        \label{ICMLDALfig:ECLAvsFR}
    \end{subfigure}
      \begin{subfigure}[b]{.245\textwidth}\includegraphics[width=\textwidth]{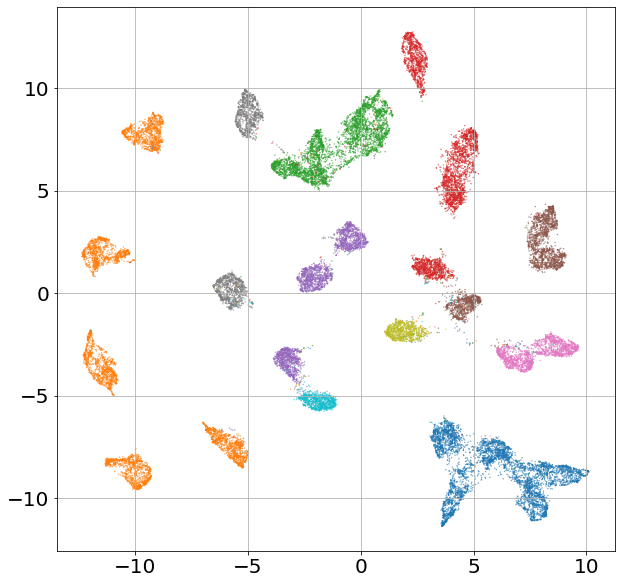}
           \centering
        \caption{ FR}
        \label{ICMLDALfig:Catfor_EWC}
    \end{subfigure}
          \begin{subfigure}[b]{.245\textwidth}\includegraphics[width=\textwidth]{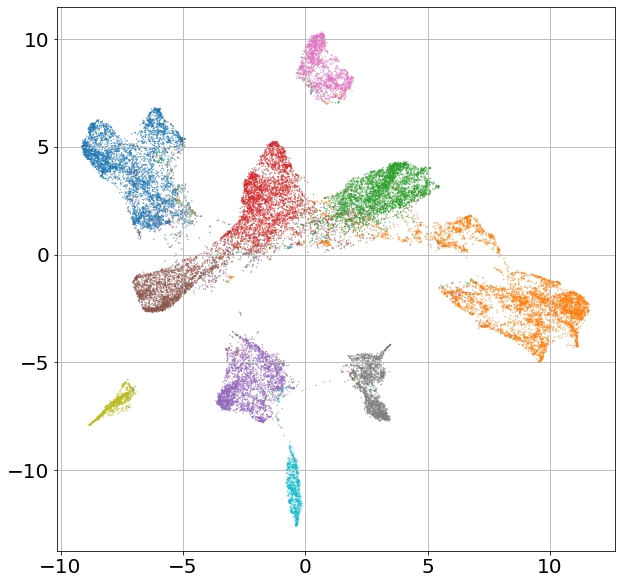}
              \centering
        \caption{ ICLA }
        \label{ICMLICMLDALfig:Catfor_Ours}
    \end{subfigure}
     \caption{Learning curves  for the five continual incremental learning tasks, designed using the permuted MNIST   tasks (a) FR (solid) vs. MB (dotted), (b) FR (solid) vs. ICLA (dotted); UMAP visualization of (c) FR and (d) ICLA  in the embedding space.  (Best viewed in color.)
     }\label{ICMLDALfig:resultsCatforgetPMNIST}
\end{figure*}
 
\subsection{Continual Incremental Learning}
 
Permuted MNIST task is a common supervised learning benchmark  for   sequential task learning~\cite{kirkpatrick2017overcoming}.  The sequential tasks are generated using  the  MNIST dataset. Each task $\Task{t}$ is generated by rearranging the pixels of all images in the dataset  using a fixed random predetermined permutation transform and keeping the labels as their original value.  As a result, we can generate many tasks that are diverse, yet   equally difficult. As a result, these tasks are suitable for performing controlled experiments.    Since no prior work has addressed incremental learning of drifting concepts, we should design a suitable set of tasks.

We    design   continual incremental learning tasks  that share common concepts using  five permuted MNSIT tasks. The first task is a binary classification of digits $0$ and $1$ for the MNIST dataset. For each subsequent task, we generate a permuted MNIST task but include only the previously seen digits plus two new digits in the   natural number order, e.g., the third task includes permuted versions of digit $0-5$. This means that at each task, new forms of all the previously learned concepts are encountered, i.e, we need to learn drifting concepts, in additional to new tasks. Hence, the model needs to expand its knowledge about the previously learned concepts while learning new concepts. We use a memory buffer with size of 30000 for MB.
Due to the nature of these tasks, we use a multi-layer perceptron (MLP) network.
Figure~\ref{ICMLDALfig:resultsCatforgetPMNIST} presents  learning curves for the five designed permuted MNIST tasks. In this figure, the learning curve for each task is illustrated with a different color and different line styles are used to distinguish the different methods (for enlarged versions, please refer to the Appendix).
At each epoch time-step, model performance is computed  as the average classification rate on the standard testing split of  the current and all the past learned tasks.  

Figure~\ref{ICMLDALfig:BPvsCLEER} presents learning curves for MB (dotted curves) and FR (solid curves). Unsurprisingly, FR leads to  almost perfect performance. We also observe MB is less effective in this   setting and catastrophic forgetting is severe for MB beyond the second task. The reason is that the concepts are more diverse in these tasks. As a result, it is more challenging to estimate the input distribution using a fixed number of stored samples that also decrease due to a fixed buffer size. We can conclude that as tasks become more complex, a larger memory buffer will be necessary which poses a challenge for MB.   Figure~\ref{ICMLDALfig:ECLAvsFR} presents learning curves for FR (solid curves) and MB (dotted  curve).
As can be seen, ICLA  is able to  learn drifting concepts incrementally. Again,   major forgetting effect for ICLA occurs as a sudden performance drop when learning a new task starts. 
This observation demonstrates that an important vulnerability for ICLA is the  structure of the   autoencoder that we build. This can be deduced from our theoretical result because an important condition for tightness of the provided bound in Lemma~1 is that we have: $\psi\approx \phi^{-1}$. 
Both our theoretical and experimental results suggest that if   can build auto-encoders   that can  generate  pseudo-data points with high quality, incremental learning can be performed    using ICLA. In other words, learning quality depends on the generative power the base network structure. 
 Finally, we also observe that as  more tasks are learned after learning a particular task, model performance on that particular task degrades more. This observation is compatible with the nervous system as memories fade out when time passes.

In addition to requiring  a memory buffer with an unlimited size, we also demonstrate that an issue for FR is inability to identify concepts across the tasks in the embedding space. We  use the UMAP~\cite{mcinnes2018umap} tool to reduce the dimensionality of the data representations in the embedding space to two for 2D data visualization. We    illustrated   the testing split of data  for all the tasks in the embedding space   $\mathcal{Z}$ in Figure~\ref{ICMLDALfig:Catfor_EWC} 
for FR and Figure~\ref{ICMLICMLDALfig:Catfor_Ours} for  ICLA when the final task is learned. In these figures,
each color corresponds to one of the digits $\{0,1,\ldots,9\}$. As expected from the learning curves, data points for  digits form separable clusters for both methods. This result verifies that the 
PDP hypothesis holds in these experiments and hence the internal distribution can be modeled using a GMM.
The important distinction between FR and ICLA is that FR has led to the generation of distinct clusters for each concept class per task. This means that each concept class has not been learned internally as one concept and FR learns each concepts as several distinct concepts across the domains.
This observation also serves as an ablative study for our method because it demonstrates  that matching distributions class-conditionally in the embedding space is necessary, as justified by the theoretical analysis.

 In figure~\ref{ICMLICMLDALfig:Catfor_Ours}, we  observe that  ten clusters for the ten observed concepts are formed when ICLA is used. This observation demonstrates that  ICLA  is able to track modes of the GMM successfully as more tasks are learned. ICLA is also able to build  concept classes that are semantically meaningful across all tasks based on the labels. This is the reason that we can learn new classes incrementally in  a continual lifelong learning scenario. In other words, as opposed to FR, ICLA encodes each cross-task concept in a single mode of the internal GMM distribution. This allows for expanding concepts for cross-domain abstraction similar to humans.

\section{Conclusions}
\label{ConclusionsUAI2021}
We developed an algorithm for  continual incremental learning of concepts  based on modeling the internal  distribution of input data as a GMM and then   updating the GMM as new experiences are acquired. We   track this distribution to accumulate the new learned knowledge   to the past learned knowledge consistently.  We expand the base classifier model to make a generative model to  allow for generating a pseudo-dataset   for pseudo-rehearsal and   experience replay. We provided theoretical   and empirical result  to validate our algorithm. 
 
 

\section*{Ethical Statement}

We foresee  no  significant ethical issues for our work.


\bibliographystyle{named}
\bibliography{ijcai23}

\begin{thebibliography}{}

\bibitem[\protect\citeauthoryear{Aljundi \bgroup \em et al.\egroup
  }{2019}]{aljundi2019gradient}
Rahaf Aljundi, Min Lin, Baptiste Goujaud, and Yoshua Bengio.
\newblock Gradient based sample selection for online continual learning.
\newblock In {\em Advances in Neural Information Processing Systems}, pages
  11816--11825, 2019.

\bibitem[\protect\citeauthoryear{Bolley \bgroup \em et al.\egroup
  }{2007}]{bolley2007quantitative}
Fran{\c{c}}ois Bolley, Arnaud Guillin, and C{\'e}dric Villani.
\newblock Quantitative concentration inequalities for empirical measures on
  non-compact spaces.
\newblock {\em Probability Theory and Related Fields}, 137(3-4):541--593, 2007.

\bibitem[\protect\citeauthoryear{Bonneel \bgroup \em et al.\egroup
  }{2015}]{bonneel2015sliced}
Nicolas Bonneel, Julien Rabin, Gabriel Peyr{\'e}, and Hanspeter Pfister.
\newblock Sliced and radon wasserstein barycenters of measures.
\newblock {\em Journal of Math. Imag. and Vision}, 51(1):22--45, 2015.

\bibitem[\protect\citeauthoryear{Chen and Liu}{2016}]{chen2016lifelong}
Zhiyuan Chen and Bing Liu.
\newblock Lifelong machine learning.
\newblock {\em Synthesis Lectures on Artificial Intelligence and Machine
  Learning}, 10(3):1--145, 2016.

\bibitem[\protect\citeauthoryear{Courty \bgroup \em et al.\egroup
  }{2017}]{courty2017optimal}
Nicolas Courty, R{\'e}mi Flamary, Devis Tuia, and Alain Rakotomamonjy.
\newblock Optimal transport for domain adaptation.
\newblock {\em IEEE TPAMI}, 39(9):1853--1865, 2017.

\bibitem[\protect\citeauthoryear{Diekelmann and Born}{2010}]{diek2010}
Susanne Diekelmann and Jan Born.
\newblock The memory function of sleep.
\newblock {\em Nature Reviews Neuroscience}, 11(2):114, 2010.

\bibitem[\protect\citeauthoryear{French}{1991}]{french1991using}
Robert~M French.
\newblock Using semi-distributed representations to overcome catastrophic
  forgetting in connectionist networks.
\newblock In {\em Proceedings of the 13th annual cognitive science society
  conference}, volume~1, pages 173--178, 1991.

\bibitem[\protect\citeauthoryear{French}{1999}]{french1999catastrophic}
Robert~M French.
\newblock Catastrophic forgetting in connectionist networks.
\newblock {\em Trends in Cog. Sciences}, 3(4):128--135, 1999.

\bibitem[\protect\citeauthoryear{Gama \bgroup \em et al.\egroup
  }{2014}]{gama2014survey}
Jo{\~a}o Gama, Indr{\.e} {\v{Z}}liobait{\.e}, Albert Bifet, Mykola Pechenizkiy,
  and Abdelhamid Bouchachia.
\newblock A survey on concept drift adaptation.
\newblock {\em CSUR}, 46(4):1--37, 2014.

\bibitem[\protect\citeauthoryear{Gennari \bgroup \em et al.\egroup
  }{1989}]{gennari1989models}
John~H Gennari, Pat Langley, and Doug Fisher.
\newblock Models of incremental concept formation.
\newblock {\em Artificial intelligence}, 40(1-3):11--61, 1989.

\bibitem[\protect\citeauthoryear{Hasson \bgroup \em et al.\egroup
  }{2020}]{hasson2020direct}
Uri Hasson, Samuel~A Nastase, and Ariel Goldstein.
\newblock Direct fit to nature: An evolutionary perspective on biological and
  artificial neural networks.
\newblock {\em Neuron}, 105(3):416--434, 2020.

\bibitem[\protect\citeauthoryear{He and Jaeger}{2018}]{he2018overcoming}
Xu~He and Herbert Jaeger.
\newblock Overcoming catastrophic interference using conceptor-aided
  backpropagation.
\newblock In {\em International Conference on Learning Representations}, 2018.

\bibitem[\protect\citeauthoryear{Hinton \bgroup \em et al.\egroup
  }{1984}]{hinton1984distributed}
Geoffrey~E Hinton, James~L McClelland, and David~E Rumelhart.
\newblock Distributed representations.
\newblock 1984.

\bibitem[\protect\citeauthoryear{Jiang and Conrath}{1997}]{jiang1997semantic}
Jay~J Jiang and David~W Conrath.
\newblock Semantic similarity based on corpus statistics and lexical taxonomy.
\newblock In {\em Proceedings of the 10th Research on Computational Linguistics
  International Conference}, pages 19--–33, 1997.

\bibitem[\protect\citeauthoryear{Kirkpatrick \bgroup \em et al.\egroup
  }{2017}]{kirkpatrick2017overcoming}
James Kirkpatrick, Razvan Pascanu, Neil Rabinowitz, and Others.
\newblock Overcoming catastrophic forgetting in neural networks.
\newblock {\em Proceedings of the national academy of sciences},
  114(13):3521--3526, 2017.

\bibitem[\protect\citeauthoryear{Lake \bgroup \em et al.\egroup
  }{2015}]{lake2015human}
Brenden~M Lake, Ruslan Salakhutdinov, and Joshua~B Tenenbaum.
\newblock Human-level concept learning through probabilistic program induction.
\newblock {\em Science}, 350(6266):1332--1338, 2015.

\bibitem[\protect\citeauthoryear{Lamprecht and
  LeDoux}{2004}]{lamprecht2004structural}
Raphael Lamprecht and Joseph LeDoux.
\newblock Structural plasticity and memory.
\newblock {\em Nature Reviews Neuroscience}, 5(1):45, 2004.

\bibitem[\protect\citeauthoryear{Lee \bgroup \em et al.\egroup
  }{2017}]{lee2017overcoming}
Sang-Woo Lee, Jin-Hwa Kim, Jaehyun Jun, Jung-Woo Ha, and Byoung-Tak Zhang.
\newblock Overcoming catastrophic forgetting by incremental moment matching.
\newblock In {\em Advances in neural information processing systems}, pages
  4652--4662, 2017.

\bibitem[\protect\citeauthoryear{Li and Hoiem}{2018}]{li2018learning}
Zhizhong Li and Derek Hoiem.
\newblock Learning without forgetting.
\newblock {\em IEEE Transactions on Pattern Analysis and Machine Intelligence},
  40(12):2935--2947, 2018.

\bibitem[\protect\citeauthoryear{Lopez-Paz and
  Ranzato}{2017}]{lopez2017gradient}
David Lopez-Paz and Marc'Aurelio Ranzato.
\newblock Gradient episodic memory for continual learning.
\newblock In {\em Advances in neural information processing systems}, pages
  6467--6476, 2017.

\bibitem[\protect\citeauthoryear{McClelland and
  Rogers}{2003}]{mcclelland2003parallel}
James~L McClelland and Timothy~T Rogers.
\newblock The parallel distributed processing approach to semantic cognition.
\newblock {\em Nature reviews Neuro.}, 4(4):310--322, 2003.

\bibitem[\protect\citeauthoryear{McClelland \bgroup \em et al.\egroup
  }{1986}]{mcclelland1986parallel}
James~L McClelland, David~E Rumelhart, PDP~Research Group, et~al.
\newblock Parallel distributed processing.
\newblock {\em Explorations in the Microstructure of Cognition}, 2:216--271,
  1986.

\bibitem[\protect\citeauthoryear{McClelland \bgroup \em et al.\egroup
  }{1995}]{mcclelland1995there}
James~L McClelland, Bruce~L McNaughton, and Randall~C O'Reilly.
\newblock Why there are complementary learning systems in the hippocampus and
  neocortex: Insights from the successes and failures of connectionist models
  of learning and memory.
\newblock {\em Psychological Review}, 102(3):419, 1995.

\bibitem[\protect\citeauthoryear{McInnes \bgroup \em et al.\egroup
  }{2018}]{mcinnes2018umap}
Leland McInnes, John Healy, and James Melville.
\newblock Umap: Uniform manifold approximation and projection for dimension
  reduction.
\newblock {\em arXiv preprint arXiv:1802.03426}, 2018.

\bibitem[\protect\citeauthoryear{Morgenstern \bgroup \em et al.\egroup
  }{2014}]{morgenstern2014properties}
Yaniv Morgenstern, Mohammad Rostami, and Dale Purves.
\newblock Properties of artificial networks evolved to contend with natural
  spectra.
\newblock {\em Proceedings of the National Academy of Sciences}, 111(Supplement
  3):10868--10872, 2014.

\bibitem[\protect\citeauthoryear{Rannen \bgroup \em et al.\egroup
  }{2017}]{rannen2017encoder}
Amal Rannen, Rahaf Aljundi, Matthew~B Blaschko, and Tinne Tuytelaars.
\newblock Encoder based lifelong learning.
\newblock In {\em Proceedings of the IEEE International Conference on Computer
  Vision}, pages 1320--1328, 2017.

\bibitem[\protect\citeauthoryear{Rasch and Born}{2013}]{rasch2013}
Bj{\"o}rn Rasch and Jan Born.
\newblock About sleep's role in memory.
\newblock {\em Physiological Reviews}, 93(2):681--766, 2013.

\bibitem[\protect\citeauthoryear{Rebuffi \bgroup \em et al.\egroup
  }{2017}]{rebuffi2017icaRL}
Sylvestre-Alvise Rebuffi, Alexander Kolesnikov, Georg Sperl, and Christoph~H
  Lampert.
\newblock icarl: Incremental classifier and representation learning.
\newblock In {\em Proceedings of the IEEE conference on Computer Vision and
  Pattern Recognition}, pages 2001--2010, 2017.

\bibitem[\protect\citeauthoryear{Redko \bgroup \em et al.\egroup
  }{2017}]{redko2017theoretical}
Ievgen Redko, Amaury Habrard, and Marc Sebban.
\newblock Theoretical analysis of domain adaptation with optimal transport.
\newblock In {\em Joint European Conference on Machine Learning and Knowledge
  Discovery in Databases}, pages 737--753. Springer, 2017.

\bibitem[\protect\citeauthoryear{Robins}{1995}]{robins1995catastrophic}
Anthony Robins.
\newblock Catastrophic forgetting, rehearsal and pseudorehearsal.
\newblock {\em Connection Science}, 7(2):123--146, 1995.

\bibitem[\protect\citeauthoryear{Rostami and
  Galstyan}{2023}]{rostami2023overcoming}
Mohammad Rostami and Aram Galstyan.
\newblock Overcoming concept shift in domainaware settings through consolidated
  internal distributions.
\newblock In {\em Proceedings of the AAAI conference on artificial
  intelligence}, volume~1, 2023.

\bibitem[\protect\citeauthoryear{Rostami \bgroup \em et al.\egroup
  }{2018}]{rostami2018crowdsourcing}
Mohammad Rostami, David Huber, and Tsai-Ching Lu.
\newblock A crowdsourcing triage algorithm for geopolitical event forecasting.
\newblock In {\em Proceedings of the 12th ACM Conference on Recommender
  Systems}, pages 377--381, 2018.

\bibitem[\protect\citeauthoryear{Rostami \bgroup \em et al.\egroup
  }{2019}]{rostami2019Complementary}
Mohammad Rostami, Soheil Kolouri, and Praveen Pilly.
\newblock Complementary learning for overcoming catastrophic forgetting using
  experience replay.
\newblock In {\em IJCAI}, 2019.

\bibitem[\protect\citeauthoryear{Rostami \bgroup \em et al.\egroup
  }{2020a}]{rostami2020using}
Mohammad Rostami, David Isele, and Eric Eaton.
\newblock Using task descriptions in lifelong machine learning for improved
  performance and zero-shot transfer.
\newblock {\em Journal of Artificial Intelligence Research}, 67:673--704, 2020.

\bibitem[\protect\citeauthoryear{Rostami \bgroup \em et al.\egroup
  }{2020b}]{rostami2020generative}
Mohammad Rostami, Soheil Kolouri, Praveen Pilly, and James McClelland.
\newblock Generative continual concept learning.
\newblock In {\em Proceedings of the AAAI Conference on Artificial
  Intelligence}, volume~34, pages 5545--5552, 2020.

\bibitem[\protect\citeauthoryear{Rostami}{2021a}]{rostami2021lifelong}
Mohammad Rostami.
\newblock Lifelong domain adaptation via consolidated internal distribution.
\newblock In {\em Proceedings of the 2021 NeurIPS Conference}, 2021.

\bibitem[\protect\citeauthoryear{Rostami}{2021b}]{rostami2021transfer}
Mohammad Rostami.
\newblock {\em Transfer Learning Through Embedding Spaces}.
\newblock CRC Press, 2021.

\bibitem[\protect\citeauthoryear{Roy \bgroup \em et al.\egroup
  }{2020}]{roy2020tree}
Deboleena Roy, Priyadarshini Panda, and Kaushik Roy.
\newblock Tree-cnn: a hierarchical deep cnn for incremental learning.
\newblock {\em Neural Networks}, 121:148--160, 2020.

\bibitem[\protect\citeauthoryear{Saitoh}{1997}]{saitoh1997integral}
Saburou Saitoh.
\newblock {\em Integral transforms, reproducing kernels and their
  applications}, volume 369.
\newblock CRC Press, 1997.

\bibitem[\protect\citeauthoryear{Sarwar \bgroup \em et al.\egroup
  }{2019}]{sarwar2019incremental}
Syed~Shakib Sarwar, Aayush Ankit, and Kaushik Roy.
\newblock Incremental learning in deep convolutional neural networks using
  partial network sharing.
\newblock {\em IEEE Access}, 2019.

\bibitem[\protect\citeauthoryear{Saxe \bgroup \em et al.\egroup
  }{2019}]{saxe2019mathematical}
Andrew~M Saxe, James~L McClelland, and Surya Ganguli.
\newblock A mathematical theory of semantic development in deep neural
  networks.
\newblock {\em Proceedings of the National Academy of Sciences}, page
  201820226, 2019.

\bibitem[\protect\citeauthoryear{Schaul \bgroup \em et al.\egroup
  }{2016}]{schaul2015prioritized}
Tom Schaul, John Quan, Ioannis Antonoglou, and David Silver.
\newblock Prioritized experience replay.
\newblock In {\em IJCLR}, 2016.

\bibitem[\protect\citeauthoryear{Shin \bgroup \em et al.\egroup
  }{2017}]{shin2017continual}
Hanul Shin, Jung~Kwon Lee, Jaehong Kim, and Jiwon Kim.
\newblock Continual learning with deep generative replay.
\newblock In {\em NeurIPS}, pages 2990--2999, 2017.

\bibitem[\protect\citeauthoryear{Song \bgroup \em et al.\egroup
  }{2000}]{song2000competitive}
Sen Song, Kenneth~D Miller, and Larry~F Abbott.
\newblock Competitive hebbian learning through spike-timing-dependent synaptic
  plasticity.
\newblock {\em Nature neuroscience}, 3(9):919--926, 2000.

\bibitem[\protect\citeauthoryear{Stan and Rostami}{2021}]{stan2021unsupervised}
Serban Stan and Mohammad Rostami.
\newblock Unsupervised model adaptation for continual semantic segmentation.
\newblock In {\em Proceedings of the AAAI Conference on Artificial
  Intelligence}, volume~35, pages 2593--2601, 2021.

\bibitem[\protect\citeauthoryear{Widmer and Kubat}{1996}]{widmer1996learning}
Gerhard Widmer and Miroslav Kubat.
\newblock Learning in the presence of concept drift and hidden contexts.
\newblock {\em Machine learning}, 23(1):69--101, 1996.

\bibitem[\protect\citeauthoryear{Wu \bgroup \em et al.\egroup
  }{2018}]{wu2018memory}
Chenshen Wu, Luis Herranz, Xialei Liu, Joost van~de Weijer, Bogdan Raducanu,
  et~al.
\newblock Memory replay gans: Learning to generate new categories without
  forgetting.
\newblock In {\em NeurIPS}, pages 5962--5972, 2018.

\bibitem[\protect\citeauthoryear{Zeng \bgroup \em et al.\egroup
  }{2019}]{zeng2019continual}
Guanxiong Zeng, Yang Chen, Bo~Cui, and Shan Yu.
\newblock Continual learning of context-dependent processing in neural
  networks.
\newblock {\em Nature Machine Intelligence}, 1(8):364--372, 2019.

\bibitem[\protect\citeauthoryear{Zenke \bgroup \em et al.\egroup
  }{2017}]{zenke2017temporal}
Friedemann Zenke, Wulfram Gerstner, and Surya Ganguli.
\newblock The temporal paradox of hebbian learning and homeostatic plasticity.
\newblock {\em Curr. opinion in neuro.}, 43:166--176, 2017.

\end{thebibliography}

\clearpage

\appendix

\section{Cognitive Modeling Background}
Our work is inspired by the ``complementary learning systems'' (CLS) theory within the ``parallel distributed processing'  (PDP) paradigm.

\subsection{Parallel Distributed Processing}

Parallel distributed processing (PDP)   approach in  cognitive science tries to explain mental phenomena using structures similar to artificial neural networks~\cite{mcclelland1986parallel} which were historically inspired by biological neurons and their parallel processing ability in   low-level structures of the nervous system. Within this framework, learning process is modeled as adjusting weights in a   network  according to various   rules such as Hebbian learning~\cite{song2000competitive}. PDP models data representations in the nervous system as distributed representations that are encoded in the neural activation functions~\cite{hinton1984distributed} which is analogous to representing data in a semantically meaningful embedding space.  Hasson et al.~\cite{hasson2020direct} argue that although evolution trains the biological neural networks blindly based on behavioral advantage, but the emerging behaviors are similar to behaviors that are observed in the artificial neural networks. They argue that both biological and artificial neural networks learn a meaningful embedding space by optimizing an objective function on densely sampled training data, i.e., empirical risk minimization.  As a result, the dimensions of the embedding space capture features that help to encode informative variations across the input data points. We have based our work on  this hypothesis. This means that when we train an artificial neural networks for classification, data representations encode input data similarity in terms of belonging to the same class. This means that we   model the data representation distribution using a multi-modal distribution.

\subsection{Complementary Learning Systems}
We rely on the Complementary Learning Systems (CLS) theory~\cite{mcclelland1995there} to prevent catastrophic forgetting both when concepts drift or when new concepts are observed. CLS theory is proposed within the PDP paradigm and hypothesizes that continual lifelong learning ability of the nervous system is a result of  a dual long- and short-term memory system.  The hippocampus acts as short-term memory and   encodes recent experiences that are used to consolidate the knowledge in the neocortex as long-term memory through offline experience replays during sleep~\cite{diek2010}. The hippocampal experience replay is more of a generative process because the input stimuli is absent during these replays. In our work the internal multimodal distribution models the neocortical consolidated knowledge. When a task is learned, this distribution is updated to incorporate the new learned knowledge to update the long-term memory. The hippocampal experience replay is modeled when the pseudo-dataset is generated to prevent catastrophic forgetting using pseudo-rehearsal. This pseudo-dataset is more representative of the recent memory, as demonstrated by both our theoretical   and empirical results.

 \section{Block Diagram of the Proposed Method}
\begin{figure*}[t!]
    \centering
    \includegraphics[width=\linewidth]{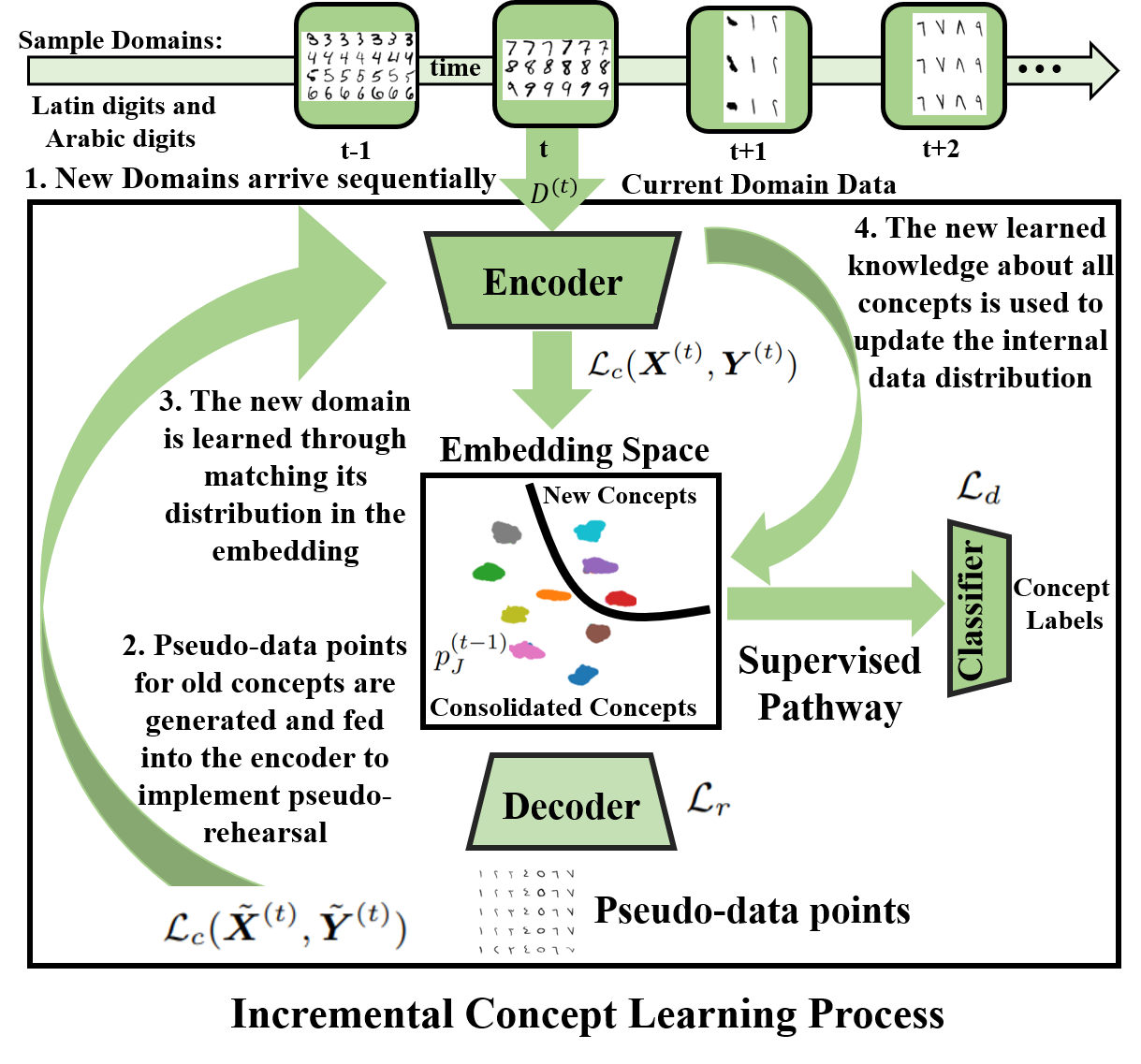}
         \caption{Block-diagram Architecture of the proposed   Incremental Learning System.  }
         \label{structureFigICMLap}
\end{figure*}

Figure~\ref{structureFigICMLap} presents an enlarged version of the system block-diagram for more clarity on how the PDP and the CLS theories are reflected in our model. 
Visualization of the data representation in the embedding space in Figure~\ref{structureFigICMLap} highlights the PDP hypothesis. An important condition for the proposed method to work is that the PDP hypothesis holds. This means that the concepts are formed as clusters in the embedding. As a result, the task data in the embedding would follow a GMM distribution and the number of components of this GMM is equal to the number of observed classes. As a result, the second term in Eq.~(3) is the distance between the empirical and the real distributions for a GMM. Hence, the second term is minimized by fitting a GMM on the  drawn distribution samples. Similar to all the parametric algorithms, our method works only if the assumption about the data distribution is correct. All parametric algorithms are limited in this sense.

After learning each task, we need to update the estimate of the internal GMM distribution in two aspects  for generating representative pseudo-dataset in the future.  First, the number of components should be updated to $k_t$ because new classes may be observed. Second, estimates for parameters of each concept cluster, i.e., the mean and the variance of the corresponding Gaussian component, should be updated to incorporate potential concept drifts. Updating this distribution models the process of knowledge consolidation in the nervous system using recent experiences.

\section{GMM Estimation}
Upon learning a task, the internal distribution  will be updated according to the input distribution.
The empirical version of the internal distribution is encoded  by the training data samples $\{(\phi_{\bm{v}}(\bm{x}_i^{(t)}),\bm{y}_i^{(t)})\}_{i=1}^{n_t}$, where with a slight abuse of notation, we use the same notation to denote the pseudo-samples. We   consider the distribution $p^{(t)}(\bm{z})$ to be a GMM with $k_t$ components: 
\begin{equation}
p^{(t)}(\bm{z})=\sum_{j=1}^{k_t} \alpha_j
\mathcal{N}(\bm{z}|\bm{\mu}_j,\bm{\Sigma}_j),
\end{equation}  
where $\alpha_j$ denotes the mixture weights, i.e., prior probability for each class, $\bm{\mu}_j$ and $\bm{\Sigma}_j$ denote the mean and co-variance for each component.
Since we have labeled data points, we can compute  the GMM parameters using MAP estimates. Let $\bm{S}_j$ denote the support set for class $j$ in the training dataset, i.e., $\bm{S}_j=\{(\bm{x}_i^{(t)},\bm{y}_i^{(t)})\in \mathcal{D}_{\mathcal{S}}|\arg\max\bm{y}_i^{(t)}=j \}$. Then, the MAP estimate for the parameters would be:  
\begin{equation}
\small
\begin{split}
&\hat{\alpha}_j = \frac{|\bm{S}_j|}{n_t},\hspace{2mm}\\&\hat{\bm{\mu}}_j = \sum_{(\bm{x}_i^{(t)},\bm{y}_i^{(t)})\in \bm{S}_j}\frac{1}{|\bm{S}_j|}\phi_v(\bm{x}_i^{(t)}),\hspace{2mm} \hat{\bm{\Sigma}}_j =\sum_{(\bm{x}_i^{(t)},\bm{y}_i^{(t)})\in \bm{S}_j}\frac{1}{|\bm{S}_j|}\big(\phi_v(\bm{x}_i^{(t)})-\hat{\bm{\mu}}_j\big)^\top\big(\phi_v(\bm{x}_i^{(t)})-\hat{\bm{\mu}}_j\big).
\end{split}
\label{eq:MAPestap}
\end{equation}    
We can use these estimates to draw samples from $\hat{p}^{(t)}(\cdot)$ to generate a representative pseudo-dataset before learning the subsequent task.

 \section{Proof of Theorem~1 and Lemma~1}
 
 Our proof is modeled after Redko et al.~\cite{redko2017theoretical}. The proof by Redko et al.~\cite{redko2017theoretical} is limited to the problem of domain adaptation in which the same classes exist across two domains. We adapt the proof to work in our learning  setting, where the two distributions share only a subset the  classes.
 
 We first review the definition of the optimal transport. Let $\Omega \subset \mathbb{R}^d$ be a measurable space and $\mathcal{P}(\Omega)$ denote the set of probability distributions that are defined over $\Omega$. Given two distributions $p(\cdot),q(\cdot)\in\mathcal{P}(\Omega)$ and the cost function $c:\Omega^2\rightarrow \mathbb{R}^+$, the optimal transport distance between $p(\cdot)$ and $q(\cdot)$ is defined as:
 \begin{equation}
\begin{split}
&W(p,q)=\underset{{\gamma \in \Pi (p,q)}}{\text{inf}}\int_{\Omega^2}c(\bm{x},\bm{y})d\gamma(\bm{x},\bm{y}),
\end{split}
\label{eq:mainSuppICML1ap}
\end{equation}   
where $\Pi(\cdot,\cdot)$ denotes the set of all joint distributions over $\Omega^2$ that have marginal distributions $p$ and $q$.   Optimal transport is well-defined for any proper selection of the cost function. In our proof, we consider that the cost function has the specif form: $c(\bm{x},\bm{y})=\|\eta(\bm{x})-\eta(\bm{y})\|_{\mathcal{G}}$, where $\eta:\mathbb{R}^d\rightarrow \mathbb{R}^{d'}$ is an embedding function and $\|\cdot\|_{\mathcal{G}}$ denotes the norm function in this space.

We will need the following lemma in our proof.

\textbf{Lemma~2}: Consider two distribution $p,q\in\mathcal{P}(\Omega)$ and two  functions $h_{\bm{w}},h_{\bm{w}'}\in\mathcal{H}$, and  the cost function $c(\bm{x},\bm{y})=\|\eta(\bm{x})-\eta(\bm{y})\|_{\mathcal{G}}$. Assume that  the hypothesis space $\mathcal{H}$ is a Reproducing Kernel Hilbert Space (RKHS) equipped with a kernel, induced by by the feature map $\eta:\Omega\rightarrow \mathbb{R}^{d'}$.
Let the loss function $\mathcal{L}(\cdot,\cdot)$ to be a mathematical metric which is convex and bounded by 1. Additionally, we assume that the loss function para metrically depends on $\|h_{\bm{w}}(\bm{x})-h_{\bm{w}'}(\bm{x})\|_{\mathcal{H}}$. Then the following inequality holds:
\begin{equation}
{\small
\begin{split}
&  \mathbb{E}_{\bm{x}\sim p}(\mathcal{L}_d(h_{\bm{w}'}(\bm{x}),h_{\bm{w}}(\bm{x}))) -\mathbb{E}_{\bm{x}\sim q}(\mathcal{L}_d(h_{\bm{w}'}(\bm{x}),h_{\bm{w}}(\bm{x})))\le W(p,q)
\end{split}}
\label{eq:mainSuppICML2ap}
\end{equation}
\textit{Proof:} First note that since the difference $h_{\bm{w}}(\bm{x})-h_{\bm{w}'}(\bm{x})$ lies in the hypothesis space, then the loss function is nonlinear function that maps a member of the $\mathcal{H}$ to positive numbers. Using results from~\cite{saitoh1997integral}, we can deduce a scalar RKHS space $\mathcal{G}$ is formed. Following the above assumptions, we can deduce:
\begin{equation}
\begin{split}
&  \mathbb{E}_{\bm{x}\sim p}(\mathcal{L}_d(h_{\bm{w}'}(\bm{x}),h_{\bm{w}}(\bm{x}))) -\mathbb{E}_{\bm{x}\sim q}(\mathcal{L}_d(h_{\bm{w}'}(\bm{x}),h_{\bm{w}}(\bm{x})))=\\&\mathbb{E}_{\bm{x}\sim p}(\langle \mathcal{L},\eta(\bm{x})-\mathbb{E}_{\bm{x}\sim q}(\langle \mathcal{L},\eta(\bm{x}) \rangle_{\mathcal{G}} )=\\& \langle\mathbb{E}_{\bm{x}\sim p}(\eta(\bm{x}))-\mathbb{E}_{\bm{x}\sim q}(\eta(\bm{x}) ) , \mathcal{L}\rangle_{\mathcal{G}}\le\\& \|\mathcal{L}\|_{\mathcal{G}}\|\mathbb{E}_{\bm{x}\sim p}(\eta(\bm{x}))-\mathbb{E}_{\bm{x}\sim q}(\eta(\bm{x}) )\|=\|\int_{\Omega} \eta d(p-q)\|_{\mathcal{G}}=\\&
\int_{\Omega^2} \|(\eta(\bm{x})-\eta(\bm{y})) d\gamma(\bm{x},\bm{y})\|_{\mathcal{G}}\le\\& \int_{\Omega^2} \|\eta(\bm{x})-\eta(\bm{y})\|_{\mathcal{G}} d\gamma(\bm{x},\bm{y})\le\\& \underset{{\gamma \in \Pi (p,q)}}{\text{inf}} \int_{\Omega^2} \|\eta(\bm{x})-\eta(\bm{y})\|_{\mathcal{G}} d\gamma(\bm{x},\bm{y})= W(p,q)
\end{split}
\label{eq:mainSuppICML31ap}
\end{equation}
 In the first and the second lines, we have used the reproducing property in $\mathcal{G}$ space. In the third and fourth lines, we first used the property of the expectation and then inner-product property. In the fifth and sixth lines, we have used the property of the joint distribution and then the definition of the optimal transport. We note that this proof is specific to a particular form of cost functions.

We also need the following result on convergence of the empirical distribution to the real distribution in the optimal transport norm in our proof.

\textbf{Theorem~2} (Theorem 1.1~\cite{bolley2007quantitative}): consider that $p(\cdot) \in\mathcal{P}(\Omega)$ and $\int_{\Omega} \exp{(\alpha \|\bm{x}\|^2_2)}dp(\bm{x})<\infty$ for some $\alpha>0$. Let $\hat{p}(\bm{x})=\frac{1}{N}\sum_i\delta(\bm{x}_i)$ denote the empirical distribution that is built from the samples $\{\bm{x}_i\}_{i=1}^N$ that are drawn i.i.d from $\bm{x}_i\sim p(\bm{x})$. Then for any $d'>d$ and $\xi<\sqrt{2}$, there exists $N_0$ such that for any $\epsilon>0$ and $N\ge N_o\max(1,\epsilon^{-(d'+2)})$, we have:
 \begin{equation}
\begin{split}
P(W(p,\hat{p})>\epsilon)\le \exp(-\frac{-\xi}{2}N\epsilon^2)
\end{split}
\label{eq:mainSuppICML3ap}
\end{equation}   

We combine the above result  and the previous lemma to prove Theorem~1.

\textbf{Theorem~1}: Consider two tasks $\Task{t}$ and $\Task{s}$ in our framework, where $s\le t$.  Let $h_{\bm{w}^{(t)}}$   be an optimal classifier trained for the $\Task{t}$ using the ICLA algorithm. Then for any $d'>d$ and $\zeta<\sqrt{2}$, there exists a constant number $N_0$ depending on $d'$ such that for any  $\xi>0$ and $\min(n_{er,t|\mathbb{N}_{s}},n_{s})\ge \max (\xi^{-(d'+2)},1)$ with probability at least $1-\xi$ for  $h_{\bm{w}^{(t)}}\in \mathcal{H}$, the following holds:
\begin{equation}
\begin{split}
e_{s}(\bm{w})\le & e_{t-1,s}(\bm{w}) +W(\hat{p}^{(t-1)}_{s}, \phi(\hat{q}^{(s)}))+e_{\mathcal{C}}(\bm{w}^*)+ \\& \sqrt{\big(2\log(\frac{1}{\xi})/\zeta\big)}\big(\sqrt{\frac{1}{n_{er,t|\mathbb{N}_{s}}}}+\sqrt{\frac{1}{n_{s}}}\big),
\end{split}
\label{eq:theroemfromcourtyCatForIncrementalap}
\end{equation}    
where $W(\cdot,\cdot)$ denotes the optimal transport distance, $n_{er,t|\mathbb{N}_{s}}$ denotes the subset of samples of the pseudo-task that belong to the classes in $\mathbb{N}_{s}$, $\phi(\hat{q}^{(s)}(\cdot))$ denotes the empirical marginal distribution for  $\Task{s}$ in the embedding space,   $\hat{p}^{(t-1)}_{s}$
denotes the conditional empirical shared distribution when the distribution $\hat{p}^{(t-1)}(\cdot)$ is conditioned to the classes in $\mathbb{N}_s$, and $e_{\mathcal{C}}(\bm{w}^*)$ denotes the optimal model for the combined risk of the two tasks on the shared classes in $\mathbb{N}_s$, i.e., $\bm{w}^*= \arg\min_{\bm{w}} e_{\mathcal{C}}(\theta)=\arg\min_{\bm{w}}\{ e_{t,s}(\bm{w})+  e_{s}(\bm{w})\}$.

\textit{Proof:} 
\begin{equation}
{\small
\begin{split}
&e_{s}(\bm{w})\le e_{s}(\bm{w}^*)+\mathbb{E}_{\bm{x}\sim \phi(q^{(s)} )}(\mathcal{L}_d(h_{\bm{w}^*}(\bm{x}),h_{\bm{w}}(\bm{x})))=\\&\Big\{e_{s}(\bm{w}^*)+\mathbb{E}_{\bm{x}\sim \phi(q^{(s)})}(\mathcal{L}_d(h_{\bm{w}^*}(\bm{x}),h_{\bm{w}}(\bm{x}))) +\\&\mathbb{E}_{\bm{x}\sim \hat{p}^{(t-1)}_{s}}(\mathcal{L}_d(h_{\bm{w}^*}(\bm{x}),h_{\bm{w}}(\bm{x})))\\&-\mathbb{E}_{\bm{x}\sim \hat{p}^{(t-1)}_{s}}(\mathcal{L}_d(h_{\bm{w}^*}(\bm{x}),h_{\bm{w}}(\bm{x})))\Big\}\le\\&
e_{s}(\bm{w}^*)+\mathbb{E}_{\bm{x}\sim \hat{p}^{(t-1)}_{s}}(\mathcal{L}_d(h_{\bm{w}^*}(\bm{x}),h_{\bm{w}}(\bm{x})))+W(p^{(t-1)}_{s}, \phi(q^{(s)}))\le\\&
e_{s}(\bm{w}^*)+e_{t-1,s}(\bm{w})+e_{t-1,s}(\bm{w}^*)+W(p^{(t-1)}_{s}, \phi(q^{(s)}))=\\&e_{t-1,s}(\bm{w})+e_{\mathcal{C}}(\bm{w}^*)+W(p^{(t-1)}_{s}, \phi(q^{(s)}))\le\\&
e^{t-1,s}(\bm{w})+e_{\mathcal{C}}(\bm{w}^*)+W(p^{(t-1)}_{s}, \hat{p}^{(t-1)}_{s})+W(\hat{p}^{(t-1)}_{s}, \phi(q^{(s)}))\le\\&
e_{t-1,s}(\bm{w})+W(\hat{p}^{(t-1)}_{s}, \phi(\hat{q}^{(s)}))+e_{\mathcal{C}}(\bm{w}^*)+\\&W(p^{(t-1)}_{s}, \hat{p}^{(t-1)}_{s})+W(\phi(\hat{q}^{(s)}), \phi(q^{(s)}))\le\\&
e_{t-1,s}(\bm{w}) +W(\hat{p}^{(t-1)}_{s}, \phi(\hat{q}^{(s)}))+e_{\mathcal{C}}(\bm{w}^*) \\& +\sqrt{\big(2\log(\frac{1}{\xi})/\zeta\big)}\big(\sqrt{\frac{1}{\tilde{n}_{t|\mathbb{N}_{s}}}}+\sqrt{\frac{1}{n_{s}}}\big)
\end{split}}
\label{eq:mainSuppICML6ap}
\end{equation}
In the above proof, fifth line is deduced from Lemma~1. In the sixth, we have used the triangular inequality on the loss function. In the seventh line, we have used the definition of the joint optimal model. In the lines eighth to tenth, we have used the triangular inequality on the optimal transport. In the last two lines, we have used Theorem~2.

We can now use Theorem~1 to deduce Lemma~1.

\textbf{Lemma~1 }: Consider the ICLA algorithm     after  learning $\Task{T}$. Then all tasks $t<T$ and under the conditions of Theorem~1, we can conclude the  following inequality:
\begin{equation}
\begin{split}
&e_{t}(\bm{w})\le  e_{T-1,t}(\bm{w})+ W(\phi(\hat{q}^{(t)}), \hat{p}_{t}^{(t)})+e_{\mathcal{C}}(\bm{w}^*)+\\
&\sum_{s=t}^{T-2} W(\hat{p}_{t}^{(s)}, \hat{p}_{t}^{(s+1)}) +\sqrt{\big(2\log(\frac{1}{\xi})/\zeta\big)}\big(\sqrt{\frac{1}{n_t}}+\sqrt{\frac{1}{\tilde{n}_{t|\mathbb{N}_t}}}\big),
\end{split}
\label{eq:theroemfromcourtyCatForoursIncrementalap}
\end{equation}    
\textit{Proof}: We consider $\Task{t}$ with empirical the distribution $\phi(\hat{q}^{(t)})$ in the embedding space   and the pseudo-task with the distribution  $\hat{p}^{(T-1)}$  in Theorem~1. Applying the triangular inequality on the term $W(\phi(\hat{q}^{(t)}), \hat{p}_{t}^{(T-1)})$ recursively, i.e., $W(\phi(\hat{q}^{(t)}), \hat{p}_{t}^{(T-1)})\le W(\phi(\hat{p}^{(t)}), \hat{p}_{t}^{(T-2)})+W( \hat{p}_{t}^{(T-2)},  \hat{p}_{t}^{(T-1)})$ for all $t \le s< T$ concludes  Lemma~1.

\section{Details of Experimental Implementation }

\subsubsection{ Datasets} 

We investigate the empirical performance of our proposed method using two commonly used benchmark datasets: MNIST ($\mathcal{M}$) and  
Fashion-MNIST ($\mathcal{U}$).   MNIST is a collection of hand written digits in $28 \times 28$ pixels with 60000 and 10000 training and testing data points, respectively. Fashion-MNIST has similar properties but the images are more realistic.     To generate permuted MNIST tasks, we followed the literature and applied a fixed random permutation to all the MNIST data points for generating each sequential task. We used cross entropy loss as the discrimination loss and the Euclidean norm as the Reconstruction loss. We used Keras for implementation and   ADAM optimizer. We run our code on a cluster node equipped with 2 Nvidia Tesla P100-SXM2 GPU's.

\subsubsection{ Evaluation Methodology}

All these datasets have their own standard testing splits. For each experiment,  we used these testing splits to measure performance of the methods that we report in terms of classification accuracy. We used  classification rate on  the  testing set of all the learned tasks to measure performance of the algorithms.   At each   training epoch, we compute the performance on the testing split of these tasks to generate the learning curves.  We performed 5 learning trials on the training sets and reported the average performance on the testing sets for these trials. 
We used brute force search to cross-validate the  parameters for each sequential task.

\subsubsection{ Network Structure}
For visual recognition experiments, we used a convolutional structure as spatial visual similarity can be captured by convolutional structures. We used a VGG16-based encoder.  The decoder subnetwork is generated by mirroring this structure. We flatten the last convolutional layer response and used a dense layer to form the embedding space with dimension 64. The classifier subnetwork is a single layer with sigmoid.

Following the literature, we have used an MLP with two layers for tasks of Table~1. The first layers has 100 nodes and the second layer has  nodes equal to the number of learned concepts.

  \begin{figure*}[t]
    \centering
           \begin{subfigure}[b]{0.45\textwidth}\includegraphics[width=\textwidth]{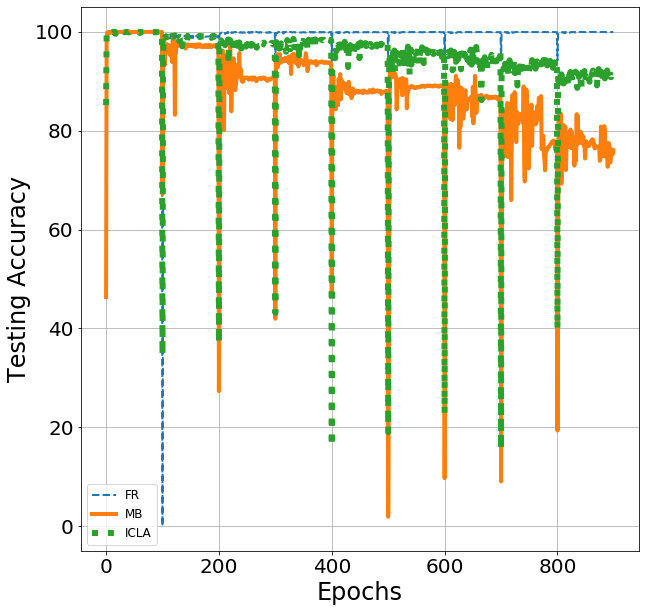}
           \centering
        \caption{MNIST}
        \label{ICMLDALfig:MNISTUSPSap}
    \end{subfigure}
    \begin{subfigure}[b]{0.45\textwidth}\includegraphics[width=\textwidth]{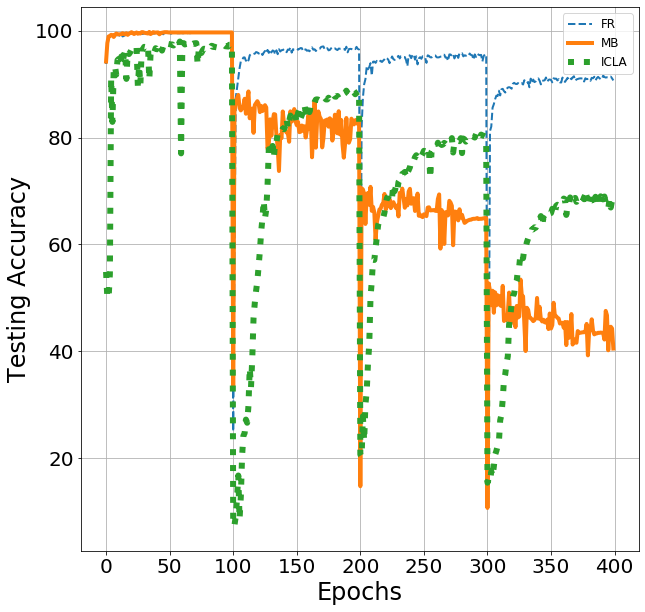}
           \centering
        \caption{FMNIST }
        \label{ICMLDALfig:USPSMNISTap}
    \end{subfigure}
     \caption{Learning curves  for the   incremental learning experiments  (a)   MNIST     and (b) Fashion-MNIST (FMNIST) datasets; (c) MNIST performance comparison (Best viewed in color on screen. Enalarged version are included in the Appendix.)  }\label{ICMLDALfig:resultsCatforgetRelatedap}
\end{figure*}

\begin{figure*}[tb!]
    \centering
           \begin{subfigure}[b]{.45\textwidth}\includegraphics[width =\textwidth,]{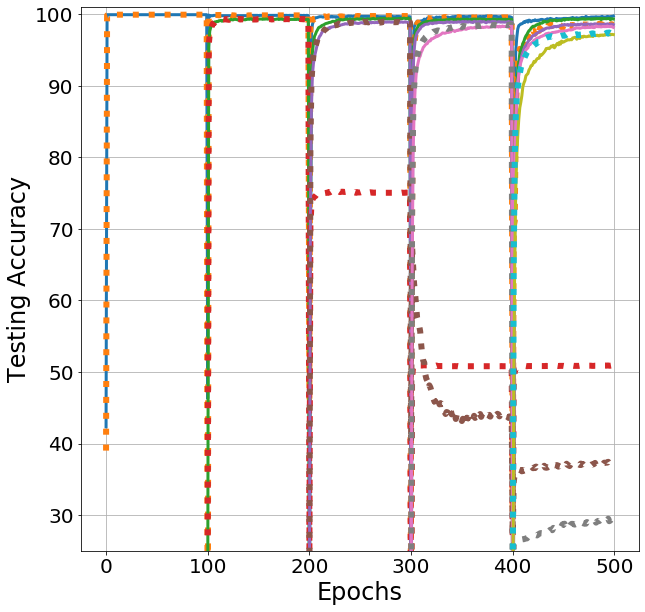}
           \centering
        \caption{FR vs. MB}
        \label{ICMLDALfig:BPvsCLEERap}
    \end{subfigure}
    \begin{subfigure}[b]{.45\textwidth}\includegraphics[width=\textwidth]{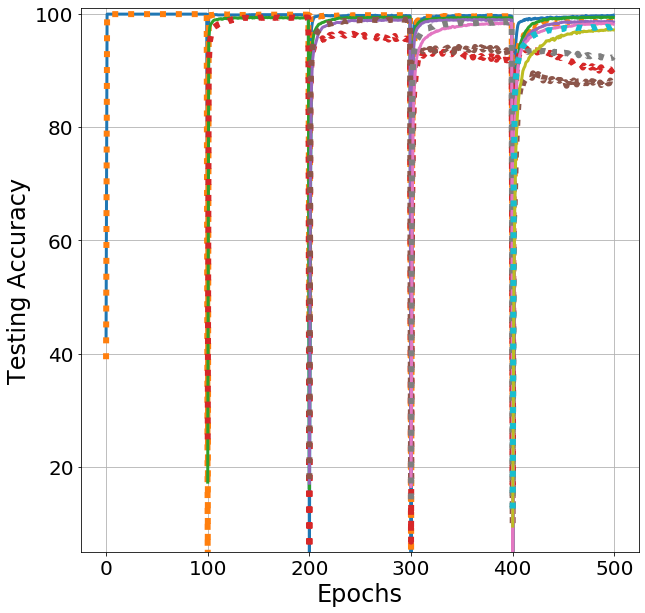}
           \centering
        \caption{FR vs. ICLA}
        \label{ICMLDALfig:ECLAvsFRap}
    \end{subfigure}
      \begin{subfigure}[b]{.45\textwidth}\includegraphics[width=\textwidth]{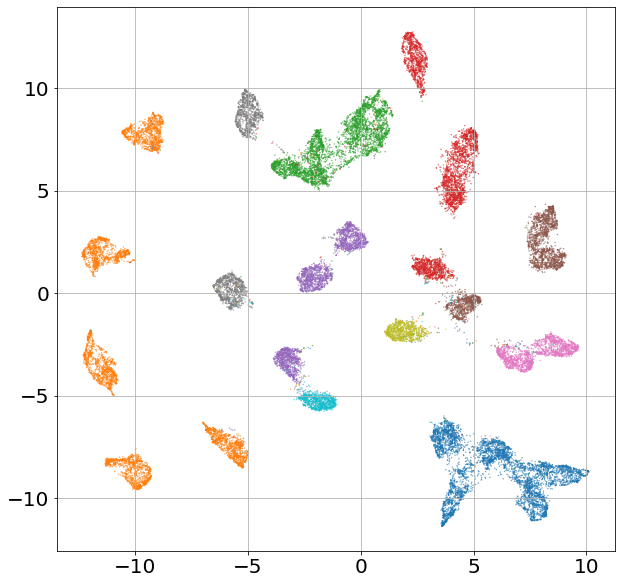}
           \centering
        \caption{ FR}
        \label{ICMLDALfig:Catfor_EWCap}
    \end{subfigure}
          \begin{subfigure}[b]{.45\textwidth}\includegraphics[width=\textwidth]{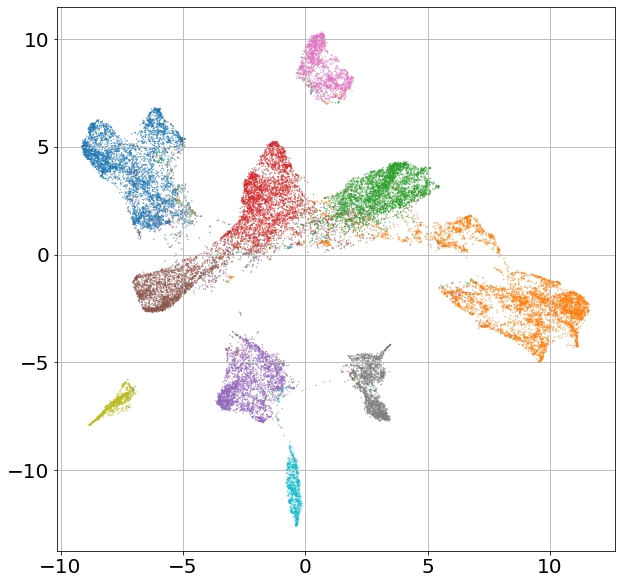}
              \centering
        \caption{ ICLA }
        \label{ICMLICMLDALfig:Catfor_Oursap}
    \end{subfigure}
     \caption{Learning curves  for the five continual incremental learning tasks, designed using the permuted MNIST   tasks (a) FR (solid) vs. MB (dotted), (b) FR (solid) vs. ICLA (dotted); UMAP visualization of (c) FR and (d) ICLA  in the embedding space.  (Best viewed in color on screen) }\label{ICMLDALfig:resultsCatforgetPMNISTap}
\end{figure*} 

 For permuted MNIST experiments, we used an  MLP network. This selection is natural as the concepts are related through permutations which can be learned with an MLP structure better. For this reason, the images were normalized and converted to $784\times 1$ vectors. The network had three hidden layers with 512, 256, and 32 nodes, respectively. We used ReLu activation between the hidden layers and selected the third hidden layer as the embedding space. This selection is natural because the last hidden layer, supposedly should respond to more abstract concepts. The decoder subnetwork is generated by mirroring the encoder subnetwork and the classifier subnetwork is a one layer with 10 nodes and sigmoid activation.

\subsection{Enlarged Figures}

For possibility of better inspection by readers, enlarged versions of Figure~2 and Figure~3 in the main body of the paper are provided in this section.

\end{document}